\documentclass[journal]{IEEEtran}
\ifCLASSINFOpdf
\else
\fi
\usepackage{xcolor}
\usepackage{amsmath,amsfonts,amssymb} 
\usepackage{mathtools} 
\usepackage{amsthm} 
\usepackage{subcaption} 
\usepackage{alltt, xspace, times, epsfig, url} 
\usepackage{gensymb,xfrac,array,booktabs,bm,xcolor} 
\usepackage[linesnumbered,lined,ruled]{algorithm2e} 
\usepackage[]{graphicx, graphics} 
\usepackage{multirow} 
\usepackage{multicol} 
\usepackage{pifont}
\usepackage{algorithmic}
\usepackage{cite}

\newtheorem{theorem}{Theorem}

\newtheorem{lemma}{Lemma}
\newtheorem{definition}{Definition}

\makeatletter
\newcommand{\nosemic}{\renewcommand{\@endalgocfline}{\relax}}
\newcommand{\dosemic}{\renewcommand{\@endalgocfline}{\algocf@endline}}
\let\oldnl\nl
\newcommand{\nonl}{\renewcommand{\nl}{\let\nl\oldnl}}

\begin{document}
%
\title{Gradient Leakage Attack Resilient Deep Learning}
%
%
%

\author{Wenqi Wei,~\IEEEmembership{Student Member,~IEEE,}
        Ling~Liu,~\IEEEmembership{Fellow,~IEEE,}
\IEEEcompsocitemizethanks{\IEEEcompsocthanksitem Wenqi Wei and Ling Liu are with School of Computer Science, Georgia Institute of Technology, Atlanta, 
GA, 30332.\protect\\ 
E-mail: wenqiwei@gatech.edu, ling.liu@cc.gatech.edu}
\thanks{Manuscript received xxxx xx, xxxx; revised xxxxx xx, xxxx.}
}

%
%

\markboth{IEEE Transactions on Information Forensics and Security,~Vol.~xx, No.~x, xxxx~20xx}%
{Shell \MakeLowercase{\textit{et al.}}: Bare Demo of IEEEtran.cls for IEEE Journals}
%



\maketitle

\begin{abstract}
Gradient leakage attacks are considered one of the wickedest privacy threats in deep learning as attackers covertly spy gradient updates during iterative training without compromising model training quality, and yet secretly reconstruct sensitive training data using leaked gradients with high attack success rate. Although deep learning with differential privacy is a defacto standard for publishing deep learning models with differential privacy guarantee, we show that differentially private algorithms with fixed privacy parameters are vulnerable against gradient leakage attacks. This paper investigates alternative approaches to gradient leakage resilient deep learning with differential privacy (DP). 
\textit{First}, we analyze existing implementation of deep learning with differential privacy, which use fixed noise variance to injects constant noise to the gradients in all layers using fixed privacy parameters.
Despite the DP guarantee provided, the method suffers from low accuracy and is vulnerable to gradient leakage attacks.
\textit{Second}, 
we present a gradient leakage resilient deep learning approach with differential privacy guarantee by using dynamic privacy parameters.  Unlike fixed-parameter strategies that result in constant noise variance, different dynamic parameter strategies present alternative techniques to introduce adaptive noise variance and adaptive noise injection which are closely aligned to the trend of gradient updates during differentially private model training. 
\textit{Finally}, 
we describe four complementary metrics to evaluate and compare alternative approaches. Extensive experiments on six benchmark datasets show that differentially private deep learning with dynamic privacy parameters outperforms the deep learning using fixed DP parameters, 
 and existing adaptive clipping approaches in all aspects: compelling accuracy performance, strong differential privacy guarantee, and high attack resilience.
\end{abstract}


%
\IEEEpeerreviewmaketitle

\section{Introduction}

\label{sec:introduction}
%
%
%
%
\IEEEPARstart{D}{eep} neural networks (DNNs) have demonstrated superior capability of learning complex tasks with high prediction accuracy. With the premium environment for model training using rich data that companies are collecting about their users, two questions remain an overwhelming challenge: (i) how can a model be trained on private collections of sensitive data so that it can be deployed safely, minimizing disclosure of sensitive training data? and (ii) can a DNN model trained with differential privacy be trusted for its outputs against privacy intrusion?

\textbf{Privacy Risks in Deep Learning.} Deep learning is vulnerable to many privacy attacks at both training phase and prediction phase, by exploiting its large capacity from the large number of model parameters sufficient for encoding the details of the individual data.
Gradient leakage attacks are the dominating privacy threats during training phase \cite{zhu2019deep,zhao2020idlg,aono2017privacy,geiping2020inverting,wei2020framework,zhu2020r,weng2020privacy,liu2021quantitative,yin2021see,qian2020what}, assuming training data is encrypted in storage and transportation. The attacker, without any prior knowledge about the learning model, can breach secrecy and confidentiality of the training data from the intermediate gradients.  There are also privacy threats in prediction phase, e.g., model inversion~\cite{fredrikson2015model,song2017machine}, attribute inference~\cite{melis2019exploiting,ganju2018property}, membership inference~\cite{shokri2017membership,nasr2018comprehensive,hui2021practical,salem2018ml,song2021systematic,zhang2021ml} and GAN-based reconstruction attack~\cite{hitaj2017deep,wang2019beyond}. These privacy concerns aggravate with the broad deployment of deep learning applications and deep learning as a service.  Although recent studies on gradient leakage attacks were in the context of federated learning~\cite{zhu2019deep,zhao2020idlg,geiping2020inverting,wei2020framework}, the attacks are high-risk threats for both centralized cloud and edge clients because the same spying process and unauthorized read can be silently employed during model training without being noticed, and the same reconstruction algorithms can be utilized
to disclose private training data from the leaked gradients. 
This paper presents risk assessment of gradient leakage attacks and presents a gradient leakage resilient deep learning approach by extending conventional deep learning with differential privacy algorithms with dynamic privacy parameter optimizations.


\textbf{Deep Learning with Differential Privacy.\/} Differentially private deep learning is the de facto standard for publishing DNN models with provable privacy guarantee~\cite{dwork2014algorithmic}: It is extremely hard to characterize the difference in output between any two models trained using two neighboring inputs differing by at most one element. In other words, by simply observing the output of a DNN model trained using the differentially private learning algorithm, one cannot tell if a single example is used in the training. \cite{abadi2016deep} is the first to implement differentially private deep learning with moments accountant method for a much tighter privacy accounting under random sampling.
To regulate the maximum influence of the model under the two neighboring inputs, conventional approaches~\cite{abadi2016deep,mcmahan2017learning,papernot2018scalable,yu2019differentially,chaudhuri2011differentially} implement differentially private deep learning by first clipping the gradients and then applying differential privacy-controlled noise to perturb the gradients before employing the stochastic gradient descent (SGD) algorithm, ensuring that each gradient descent step is differentially private. Based on composition properties of differential privacy algorithms~\cite{dwork2014algorithmic}, the final model produced upon completion of the total training steps provides a certain level of differential privacy.


\textbf{Inherent Limitations of Baseline.\/} Inspired by the pioneer work~\cite{abadi2016deep}, many proposals in the literature~\cite{mcmahan2017learning,yu2019differentially,geyer2017differentially} and open source community~\cite{tfdldpimplementation,torchdldpimplementation} employ the fixed privacy parameter strategy to decide clipping method, define the sensitivity of gradient updates and noise scale, which results in constant noise injection throughout every step of the entire training process. Although such a rigid setting of privacy parameters has shown reasonable accuracy while providing a certain level of differential privacy guarantee, they suffer some inherent limitations. First, fixed privacy parameters induce constant differential privacy noise throughout the training, which deems unnecessary especially at the later stage of training, and leads to low accuracy utility. Second, even with differential privacy guarantee, such algorithms require careful privacy parameter selection as otherwise remaining vulnerable against gradient leakage induced privacy violation, breaking the intended privacy protection. 
Although some development aim to improve the accuracy of differentially private deep learning by optimizing the noise~\cite{abadi2016deep,bassily2014private,phan2017adaptive,papernot2018scalable,xu2020adaptive,van2018three,xiang2019differentially}, all these methods are not designated to defend against the gradient leakage attacks.


\textbf{Scope and Contributions.\/}
In this paper, we investigate alternative approaches to differentially private deep learning, aiming for strong privacy, high accuracy performance as well as high resilience against gradient leakage attacks. This paper makes three original contributions. 
{\em First}, we analyze existing implementation of deep learning with differential privacy which injects differential privacy controlled noise to perturb the gradients in all layers using fixed privacy parameters.
We show that despite the differential privacy guarantee provided, the deep learning model suffers from low accuracy utility and can be vulnerable to gradient leakage attacks.
This motivates us to analyze the inherent limitations of using fixed-parameter strategies in deep learning with differential privacy. 
{\it Second},  
we propose a differentially private deep learning approach with adaptive DP parameters to address the inherent limitations of conventional deep learning algorithms with fixed DP parameters. We propose three adaptive DP parameter optimizations to allow dynamic DP controlled noise variance such that more noise is injected in early rounds and smaller noise is used to perturb the gradients in the later rounds, including dynamic gradient clipping method, dynamic sensitivity and dynamic noise scale. We show that our approach with dynamic DP parameters can achieve high resilience against gradient leakage attacks, competitive accuracy performance, better differential privacy guarantee. Unlike fixed-parameter strategies that result in constant noise variance, differentially private deep learning with adaptive DP parameters can align noise injection to the trend of gradient updates during differentially private model training. 
\textit{Third}, 
we present four complementary metrics for evaluation and comparative analysis of alternative approaches to differentially private deep learning: (i) comparing accuracy (utility) performance under the same privacy budget, (ii) comparing privacy cost and the level of privacy protection under the same target model accuracy, and (iii) comparing resilience against gradient leakage attacks.
Extensive experiments are conducted on six benchmark datasets with five privacy accounting methods.
The results show that deep learning with dynamic differential privacy optimization on sensitivity and noise scale outperforms both the baseline and other alternative approaches with dynamic parameter optimizations, offering compelling accuracy performance, strong differential privacy guarantee, and high attack resilience.


\section{Gradient Leakage Threat Model}


Gradient leakage attacks are most relevant privacy threats in the context of deep learning, in which a curious or malicious insider may conduct unauthorized read on the gradients and reconstruct private training data based on the gradients obtained through spying over the layer-wise gradients utilized by per-step SGD in each iteration of the training. Such insider threats do not directly compromise the accuracy of the trained model and thus are much harder to detect and mitigate through reactive defense methods. We argue that effective approaches to deep learning with differential privacy can build one of the best defense methods against such threats. 
Our threat model makes the following assumptions about the insider adversary: (1) the insider adversary cannot gain access to the encrypted training data prior to training; (2) the insider adversary has no intention of compromise the training procedure or the quality of the trained model; and (3) the insider adversary may gain access to the intermediate model training parameters that are often saved as checkpoint data to allow the resume of iterative training from a given step.
Unlike white-box adversarial example attacks~\cite{szegedy2013intriguing,papernot2016limitations,carlini2017towards,wei2020robust}, gradient leakage attacks do not need any prior knowledge of the DNN training algorithm and simply use independent reconstruction algorithms on leaked gradients to infer and disclose the private training data~\cite{wei2020framework} while keeping the integrity of the training.

 \begin{figure}[t]
\centerline{\includegraphics[scale=.63]{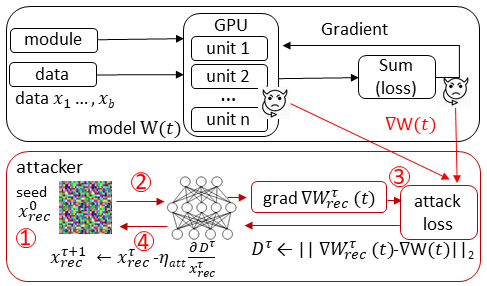}}
\vspace{-0.2cm}
\caption{\small Attack schema.}
\label{fig:attack_schema}
\vspace{-0.2cm}
\end{figure}

 \begin{figure}[t]
\centerline{\includegraphics[scale=.44]{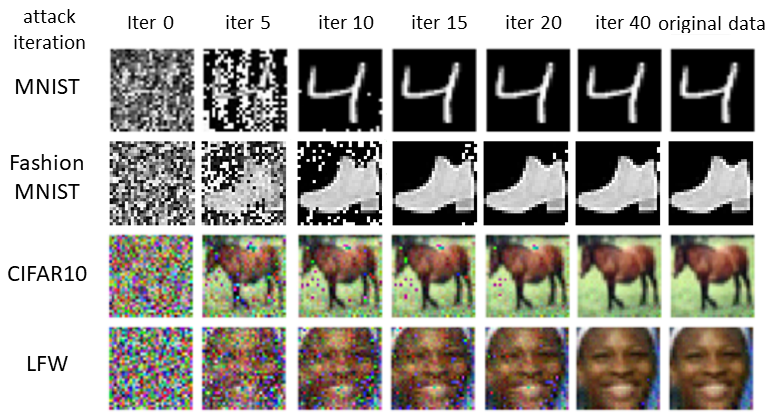}}
\vspace{-0.2cm}
\caption{\small Attack visualization.}
\label{fig:attack_vis}
\vspace{-0.4cm}
\end{figure}


With parallel processing of the data, the module is replicated on each device in the forward pass and each replica handles a portion of the input. During the backward pass, gradients from each replica are summed into the original module.
\textbf{Figure~\ref{fig:attack_schema}} gives a sketch of the gradient leakage attack algorithm under the multi-GPU setting, which configures and executes the reconstruction attack in five steps:
(1) It configures the initialization seed ($x^{0}_{rec}(t)$), a dummy data of the same resolution (or attribute structure for text) as the training data. \cite{wei2020framework} showed some significant impact of 
different initialization seeds on the attack success rate and attack cost ($\#$ attack iterations to succeed). (2) The dummy attack seed is fed into the current copy of the model. (3) The gradient of the dummy attack seed is obtained by backpropagation. (4) The gradient loss is computed using a vector distance loss function, e.g., $L_2$, between the gradient of the attack seed and the actual gradient from the training. The choice of this reconstruction loss function is another tunable attack parameter. (5) The dummy attack seed is modified by the attack reconstruction learning algorithm. It aims to minimize the vector distance loss by a loss optimizer such that the gradients of the reconstructed seed $x^{\tau}_{rec}(t)$ at round $i$ will be closer to the actual gradient updates stolen upon the completion of computing the gradients of one input in the batch  or of the entire batch. For batch gradient, \cite{zhu2019deep} and \cite{geiping2020inverting} maliciously introduce separate weights for each batch
example, making it possible for a recovery of a batch of data: $x^{\tau+1}_{rec}(t)_{i \: mod \: B} \leftarrow x^{\tau}_{rec}(t)_{i \: mod \: B} - \nabla_{x^{\tau}_{rec}(t)_{i \: mod \: B}}D $ where $B$ is the batch size and $i$ is index of data in the batch.
This attack reconstruction learning iterates until it reaches the attack termination condition ($\mathbb{T}$), typically defined by the $\#$ attack iterations, e.g., $\mathbb{T}=300$ (also a configurable attack parameter). If the reconstruction loss is smaller than the specified distance threshold then the reconstruction attack is successful. 
\textbf{Figure~\ref{fig:attack_vis}} provides a visualization by examples from four datasets at different attack iterations. The details of these datasets are given in Section~\ref{sec:experiment_utilityprivacy}. 
All the experiments on gradient leakages in this paper use the patterned random seed initialization\footnote{\url{https://github.com/git-disl/ESORICS20-CPL}}, with a $L_2$ based loss function and L-BFGS optimizer, for high attack success rate (ASR) and fast attack convergence. 





\section{Baseline Solution with Differential Privacy}
In this section, we first review the differential privacy concept.
Then we present the baseline deep learning with differential privacy approach that utilizes fixed privacy parameters and point out the inherent problem of fixed privacy parameters in terms of accuracy utility loss and gradient leakage vulnerability.

\subsection{Preliminary}
\label{sec:preliminary}

\begin{definition}\textbf{Differential privacy~\cite{dwork2014algorithmic}}: Let $\mathcal{D}$ be the domain of possible input data and $\mathcal{R}$ be the range of all possible output. A randomized mechanism $\mathcal{M}$: $\mathcal{D}\rightarrow \mathcal{R}$ satisfies ($\epsilon,\delta$)-differential privacy if for any two input sets $A \subseteq \mathcal{D}$ and $A'\subseteq \mathcal{D}$, differing with only one entry: $||A-A'||_{0}=1$, Equation~\ref{equa:dp} holds with $0 \leq \delta < 1$ and $\epsilon>0$.
\vspace{-0.1cm}
\begin{equation}
\Pr(\mathcal{M}(A) \in \mathcal{R}) \le e^{\epsilon}\Pr(\mathcal{M}(A') \in \mathcal{R}) + \delta.
\label{equa:dp}
\end{equation}
\label{def:dp}
\vspace{-0.5cm}
\end{definition}

This definition states that given $\delta$, a smaller $\epsilon$ would indicate a smaller difference between the output of $\mathcal{M}(A)$ and the output of $\mathcal{M}(A')$. By Lemma 3.17 of~\cite{dwork2014algorithmic}, $(\epsilon,\delta)$-differential
privacy ensures that for any adjacent $A, A'$, the absolute value of the privacy loss will be bounded by $\epsilon$ with probability of at least $1-\delta$. When $0 \leq \delta<1$, this definition implies that it is most likely that the observed output for $A'$ will be similar to the output for its neighboring input $A$, whereas when $\delta=0$, it indicates that the output observed under $A'$ is highly likely to be observed under $A$. Since $\delta$ is the upper bound probability of $\mathcal{M}(A)$ for breaking $\epsilon$-differential privacy, a smaller $\delta$ is desired. Following the literature~\cite{abadi2016deep,yu2019differentially,xiang2019differentially,papernot2018scalable}, $\delta$ is set to $1e-5$ in our experiments such that $\epsilon$-differential privacy would hold with probability of at least 0.9999. 


\begin{definition}
\textbf{Sensitivity~\cite{dwork2006calibrating}}: Let $\mathcal{D}$ be the domain of possible input data and $\mathcal{R}$ be the domain of all possible output.  The sensitivity of a function $f: \mathcal{D} \rightarrow \mathcal{R}$ is the maximum amount that the function value varies when a single entry of the input is changed. 
\vspace{-0.1cm}
\begin{equation}
S=\max \nolimits_{A,A' \subseteq \mathcal{D},
||A-A'||_{0}=1} ||f(A)-f(A')||_p.
\label{equa:sensitivity}
\end{equation}
\label{def:sensitivity}
\vspace{-0.5cm}
\end{definition}
\noindent
Definition~\ref{def:sensitivity} implies that to produce a randomized differentially private algorithm $\mathcal{M}(f)$ by injecting noise that follows some randomization distribution while preserving the utility of $f$, we need to bound the noise by the maximum change defined as the sensitivity of function $f$ with neighboring inputs. In this paper, we consider Gaussian mechanism as our randomization noise distribution, which adds Gaussian noise calibrated to the sensitivity of the function $f$ in $l_2$ norm. Hence, we define $S$ by $l_2$ sensitivity under Gaussian noise injection.

\begin{theorem}
\textbf{Gaussian mechanism~\cite{dwork2014algorithmic}}: Let $\mathcal{D}$ be the domain of possible input data and $\mathcal{R}$ be the range of all possible output.  With privacy parameter $\epsilon$, applying Gaussian noise $\mathcal{N}(0, \varsigma {^2})$ calibrated to a real valued function: $f: \mathcal{D}\rightarrow \mathcal{R}$ with noise variance $\varsigma {^2}$ such that $\mathcal{M}(A) = f(A) + \mathcal{N}(0, \varsigma^2)$ is $(\epsilon,\delta)$-differentially private if $\varsigma {^2} > \frac{{2\log (1.25/\delta ) \cdot {S^2}}}{{{\epsilon ^2}}}$. 
\label{theorem:gaussian}
\end{theorem}
\noindent This theorem indicates that for a constant $c$, when $c^2>2\log(1.25/\delta)$, we have Gaussian mechanism $\mathcal{M}(A) = f(A) + \mathcal{N}(0, \varsigma^2)$, and if $\varsigma \geq c S/\epsilon$, then $\left| {\log\frac{{{e^{( - 1/2{\varsigma ^2}\cdot{x^2})}}}}{{{e^{( - 1/2{\varsigma ^2}\cdot{{(x + a)}^2})}}}}} \right| \le \varepsilon $ holds with probability at least $1-\delta$, where $x$ is the random variable utilizing the Gaussian distribution noise with Gaussian density function $\frac{1}{\varsigma\sqrt{2\pi}}e^{-\frac{x^2}{2\varsigma^2}}$. Hence, it is straightforward to get the following lemma:

\begin{lemma}
Let noise variance $\varsigma^2$ in Gaussian mechanism be $\sigma^2S^2$ where $\sigma$ is the noise scale and $S$ is the $l_2$ sensitivity. We have the noise scale $\sigma$ satisfying $\sigma^2>\frac{{2\log (1.25/\delta )}}{{{\epsilon ^2}}}.$
\label{lemma:gaussian_mechanism}
\end{lemma}

According to this Lemma, noise scale $\sigma$ and privacy loss $\epsilon$ have an inverse correlation given a fixed $\delta$, i.e.,  a large noise scale indicates a small $\epsilon$, and conversely, a small noise scale implies the spending of a large privacy budget $\epsilon$. 


\subsection{Privacy Accounting}




Based on the DP composition theory~\cite{dwork2014algorithmic}, post processing theory~\cite{dwork2014algorithmic},  and privacy amplification~\cite{balle2018privacy}, the privacy budget spending can be tracked throughout the iterations of DNN training. There are four representative privacy accounting methods: moments accountant (MA)~\cite{abadi2016deep}, zCDP~\cite{yu2019differentially}, advanced composition (AdvC)~\cite{dwork2010boosting} and optimal composition (OptC)~\cite{kairouz2015composition}. 



\noindent
\begin{theorem}
\textbf{Composition theorem~\cite{dwork2014algorithmic}}:  Let $\mathcal{M}_i: \mathcal{D}\rightarrow \mathcal{R}_i$ be a randomized function that is $(\epsilon_i,\delta_i)$-differentially private. If $\mathcal{M}$ is a sequence of consecutive invocations (executions) 
of ($\epsilon_i,\delta_i$)-differentially private algorithm $\mathcal{M}_i$, then $\mathcal{M}$ is ($\sum_i \epsilon_i, \sum_i \delta_i$)-differentially private.
\label{theorem:composition_sequential}
\end{theorem}

By Theorem~\ref{theorem:composition_sequential}, a randomized function $\mathcal{M}$ that consists of a sequence of $n$ differentially private mechanisms is differentially private, and its privacy guarantee is determined by the sum of the $n$ individual privacy losses, defined by $\sum \nolimits_i^n \epsilon_i$. 
In our experiments, we consider five privacy accounting rules: base composition, advanced composition, optimal composition, Moments Accountant, and zCDP.

\textbf{Advanced composition (AdvC~\cite{dwork2010boosting}):} Let $\epsilon,\delta' \geq 0$. The class of $\epsilon_t$-differentially private mechanisms satisfies ($\epsilon_t,\delta'$)-differential
privacy under $T$-fold composition for:
\begin{equation}
    \epsilon' = \sum \nolimits_T \frac{(e^{\epsilon_{t}}-1)\epsilon_{t}}{e^{\epsilon_{t}}+1} + \sqrt{\sum \nolimits_T 2 \epsilon_{t}^2 \log(1/\delta)} 
    \label{equa:advc}
\end{equation}

\textbf{Optimal composition (OptC~\cite{kairouz2015composition}):} Let $\epsilon,\delta' \geq 0$. The class of $\epsilon_t$-differentially private mechanisms satisfies ($\epsilon_t,\delta'$)-differential
privacy under $T$-fold composition for:
\begin{equation}
    \epsilon' = \sum \nolimits_T \frac{(e^{\epsilon_{t}}-1)\epsilon_{t}}{e^{\epsilon_{t}}+1} + \sqrt{\sum \nolimits_T 2 \epsilon_{t}^2 \log(e+\frac{\sqrt{\sum\nolimits_T \epsilon_{t}^2 }}{\delta}) }
    \label{equa:optc}
\end{equation}

\textbf{Moments accountant (MA~\cite{abadi2016deep}):} For a mechanism $\mathcal{M}$ with Gaussian noise $\mathcal{N}(0,\sigma^2S^2)$ added at each iteration where $S$ denotes the sensitivity and $\sigma$ is the noise scale, given dataset $\mathcal{D}$ with a total of $N$ data points, the random sampling with replacement and the sampling rate $q = n/N$ where $n$ is the sample size, and the number of iterations $T$, if $q<\frac{1}{16\sigma}$, there exist constants $c1$ and $c2$ so that for any $\epsilon  < c1q^2T$, $\mathcal{M}$ is $(\epsilon',\delta)$-differentially private for $\delta>0$ and 
\begin{equation}
    \vspace{-0.1cm}
     \epsilon' \geq c2\frac{q\sqrt{T\log(1/\delta)}}{\sigma}
    \label{equa:momentsaccountant}
\end{equation}
Note that  R{\'{e}}nyi differential privacy~\cite{mironov2017renyi,wang2019subsampled} is proposed on top of moments accountant to keep track of the $\epsilon$ privacy spending of a sequence of randomized mechanisms with elegant composition rules. R{\'{e}}nyi differential privacy can be transformed to the standard differential privacy definition.
 We will use Moments accountant for fair comparison with other privacy accounting methods. 

\textbf{zCDP~\cite{yu2019differentially}:}  For a mechanism $\mathcal{M}$ with Gaussian noise $\mathcal{N}(0,\sigma_t^2S^2)$ added at each iteration where $S$ denotes the sensitivity and $\sigma_t$ is the per-iteration noise scale, given dataset $\mathcal{D}$ with a total of $N$ data points, given the random sampling with replacement and the sampling rate $q = n/N$ where $n$ is the sample size, and the number of iterations $T$,
if $q<\frac{1}{16\sigma}$, $\mathcal{M}$ is $(\epsilon',\delta)$-differentially private for $\delta>0$ and 
\begin{equation}
    \vspace{-0.1cm}
    \epsilon' = \sum \nolimits_T q^2/\sigma_t^2 +2\sqrt{\sum \nolimits_T q^2/\sigma_t^2\log(1/\delta)}
    \label{equa:zcdp}
\end{equation}

Due to the random sampling involved, both advanced composition and optimal composition $(\epsilon_t,\delta_t)$ consider an amplified per-iteration privacy spending and assume a fixed $\delta$ to track $\epsilon$ using the following privacy amplification theorem. Note that for Moments accountant, we follow the implementation provided by the Tensorflow Privacy module~\cite{momentsimplementation} as no closed-form expression in terms of the noise scale nor the heterogeneous per-step $\epsilon_t$ is given to compute the accumulated $\epsilon$.

\noindent
\begin{theorem}
\textbf{Privacy amplification~\cite{balle2018privacy}}: Given dataset $\mathcal{D}$ with $|D|$ data points, subsampling is defined as random sampling with replacement with $n$ as the sample size. If $\mathcal{M}$ is $(\epsilon,\delta)$-differentially private, then the subsampled mechanism with sampling rate $q=n/|D|$ is $(\log(1+q(\exp(\epsilon)-1)),q\delta)$-differentially private.
\label{theorem:amplification}
\end{theorem}



\subsection{Baseline with Fixed Parameters}    
\label{sec:dpdl}

The goal of deep learning with differential privacy is to train a $(\epsilon,\delta)$-differentially private model over $T$ iterations. By composition theorem, we need to ensure that the per-step SGD, 
denoted by $f_t$, is $(\epsilon_t,\delta_t)$-differentially private ($1\leq t \leq T$) with $\epsilon=\sum_t^T \epsilon_t$ and $\delta=\sum_t^T \delta_t$. Hence, for each iteration $t$, we inject differential privacy controlled noise to the gradients before performing per-step SGD. Given a DNN of $M$ layers, the baseline implementation for ensuring that $f_t$ is $(\epsilon_t,\delta_t)$-differentially private injects noise to all layers of the model
during each training iteration.


{\bf Fixed privacy parameters. \/} Most of existing approaches to deep learning with differential privacy~\cite{abadi2016deep,yu2019differentially,xiang2019differentially,truex2019hybrid}, including Tensorflow privacy module \cite{tfdldpimplementation} and Pytorch Opacus privacy module \cite{torchdldpimplementation}, all employ fixed privacy parameter strategies, such as the constant clipping method with pre-defined clipping bound (e.g., $C=4$), the fixed pre-defined noise scale (e.g., $\sigma=6$), and the fixed sensitivity $S$ defined using the constant clipping bound $C$. As a result, in addition to distributing the privacy budget $\epsilon$ uniformly across the total $T$ training iterations, the noise variance is fixed during the training, resulting in injecting constant noise to the gradients in each training iteration.


{\bf Constant clipping method with fixed clipping bound.\/} 
Given a training example $i$, let $\nabla W(t)_{im}$ denote the layer-wise per-example gradient vector for the $m^{th}$ layer ($1\leq m\leq M$). The clipping method is used prior to noise injection to address the problem of gradient explosion~\cite{pascanu2013difficulty}. With a constant clipping method using a fixed clipping bound $C$, the layer-wise per-example gradient vector $\nabla W(t)_{im}$ is preserved if its $l_2$ norm satisfies $||\nabla W(t)_{im}||_2 \leq C$. Otherwise, $||\nabla W(t)_{im}||_2>C$ holds, and the gradient vector $\nabla W(t)_{im}$ needs to be brought down so that its $l_2$ norm is capped by $C$. This is done by multiplying every coordinate of the gradients with a scaling factor: $C/||\nabla W(t)_{im}||_2$. 
Such per-example gradient clipping will be performed on all $M$ layers for each example $i$ in the batch $B$ of the given iteration. Let $\overline \nabla W(t)_{im}$ denote the clipped per-example gradient for layer $m$ of training example $i$ at iteration $t$ ($m\in\{1,2,...M\}$). The clipped per-example gradients are then gathered for batch averaging: $\overline \nabla W(t)=\frac{1}{B}\sum \nolimits_i^B \overline \nabla W(t)_{i}$. \textbf{Algorithm~\ref{algo:clipping_all}}
present pseudo code for baseline clipping function using the constant clipping method initialized with a preset clipping bound $C$~\cite{abadi2016deep}. 

{\bf Noise injection with fixed sensitivity $S$ and fixed $\sigma$.\/} Next, the $(\epsilon_t,\delta_t)$-differential privacy controlled Gaussian noise $\mathcal{N}(0,\sigma^2S^2)$ is injected to each layer of the batch gradients: $\widetilde \nabla W(t) = \overline \nabla W(t) + \mathcal{N}(0,\sigma^2S^2)$. The per-step SGD function $f_t$ performs the gradient descent at iteration $t$ using the Gaussian noise perturbed gradients $\widetilde \nabla W(t)$, such that $W(t+1)=W(t)-\eta \widetilde \nabla W(t)$. This process of sampling, computing gradients, clipping, and noise injection repeats until reaching the $T$ total iterations.


\begin{algorithm}[t]
\footnotesize
\caption{\footnotesize CLIP ($C$, $\nabla W_{i}$)}\label{algo:clipping_all}
\KwIn{ Per-example gradient $\nabla W_{i}$ with $M$ layers, clipping bound $C$.}
     \For{layer $m$ in $\{1,...M\}$ }{ 
      \nonl    \textbf{// compute $l_2$ norm for layer $m$ of sample $i$.} \\
        $l2_{im} = ||\nabla  W_{im}||_2$ \\
      \nonl   \textbf{// clip per-example gradients by coordinate for layer $m$ of sample $i$.}  \\
        $\overline \nabla W_{im} \leftarrow \nabla W_{im} * \min \{1, \frac{C}{l2_{im}}\}$  \label{algo:clippingline} 
         } 
\textbf{Output:} the clipped per-example gradients for sample $i$:  $\overline \nabla W_i \leftarrow \{\overline \nabla W_{im}\} , m=1,\dots, M$.
\end{algorithm}

{\bf Limitation with using Fixed Privacy Parameters. \/} 
First, using a fixed clipping bound to define the sensitivity of gradient changes for all iterations can be problematic, especially for the later iterations of training. This is because the $l_2$ norm of the layer-wise per-example gradient vector $\nabla W(t)_{im}$ satisfies $||\nabla W(t)_{im}||_2 \leq C$ in most cases as the training approaching the end. Second, with fixed sensitivity $S$ defined by the fixed clipping bound $C$, the constant noise computed based on $S$ and fixed $\sigma$ can be much larger than $C$ for $\sigma>1$. Third, injecting such large constant noise to gradients in each iteration of the training may have a detrimental effect on the accuracy performance and slow down the convergence of training and sadly it does not gain any additional privacy protection, because the accumulated privacy spending $\epsilon$ is only inversely correlated with $\sigma$.

\begin{figure}[t]
\vspace{-0.4cm}
\centerline{\includegraphics[scale=.68]{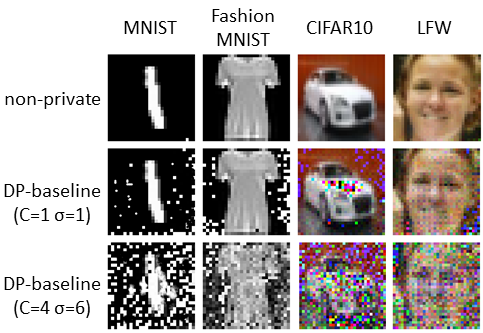}}
\vspace{-0.2cm}
\caption{\small  Illustration of gradient leakage attack resiliency for DP-baseline with different fixed privacy parameter settings.}
\label{figure:limitation_resiliency} 
\vspace{-0.4cm}
\end{figure}

Furthermore, algorithms with fixed privacy parameters may be more vulnerable to gradient leakage attacks. Any inappropriate setting of fixed privacy parameters will bring in vulnerability to gradient leakage attacks. \textbf{Figure~\ref{figure:limitation_resiliency}} provides four visual examples on four datasets: MNIST, Fashion-MNIST, LFW, and CIFAR10 under three scenarios: non-private, DP-baseline with fixed parameter setting of clipping bound $C=S=1$ and noise scale $\sigma=1$ as in~\cite{nasr2020improving,geyer2017differentially} and DP-baseline with clipping bound $C=S=4$ and noise scale $\sigma=6$ as in~\cite{abadi2016deep,yu2019differentially}. 
It is observed that an adequate amount of noise is necessary to mask the gradients from malicious or curious inference attackers using the reconstruction learning on the leaked gradients.

Given that gradients at early training iterations tend to leak more information than gradients in the later stage of the training~\cite{wei2020framework}, an obvious approach is to design a differential privacy algorithm that can inject larger noise at the early stage of training and resort to smaller noise injection as the training is close to the end. Given that the noise variance $\varsigma$ is the product of sensitivity $S$ and noise scale $\sigma$, several possible strategies can be promising, such as having the sensitivity calibrated to the $l_2$-norm of the gradients, or having a smoothly decaying noise scale such that the noise variance follows the trend of gradient updates across the $T$ training iterations.

\section{Dynamic Privacy Parameters}

\label{sec:adaptive}

A detrimental limitation of using fixed privacy parameter strategies for deciding the clipping method and for defining sensitivity and noise scale is the result of constant noise variance and constant noise injection in each iteration of deep learning, which is a root cause for poor resilience against gradient leakage attacks. In this section, we address such inherent limitations by developing dynamic parameter strategies for determining and configuring these privacy parameters in a training progress-aware manner. This brings out a significant advantage: smaller noise variance will be used to inject noise at the later stage of the training, improving the convergence speed of training with high accuracy performance; and at the same time larger noise variance will be used to 
inject a larger amount of noise at the early stage of the training, leading to higher resilience against gradient leakage attacks. Given that the noise variance $\varsigma$ is the product of sensitivity $S$ and noise scale $\sigma$ (recall Theorem~1), several possible dynamic strategies are promising, such as having the sensitivity calibrated to the $l_2$-norm of the gradients or having a smoothly decaying noise scale such that the noise variance follows the trend of gradient updates across the $T$ training iterations.

\subsection{Dynamic Sensitivity}
We describe three different strategies for implementing dynamic sensitivity, aiming to derive declining noise variance as the training progresses in iterations:  $l_2$-norm based sensitivity, denoted by DP-dynS[$l_2$-max], dynamic decaying clipping method, denoted by DP-dynS[$C_{decay}$], and the combination of both, denoted by DP-dynS. 

{\bf DP-dynS[$l_2$-max].\/} Given that the clipping with a fixed bound $C$ is performed on the $l_2$-norm of the layer-wise gradients of per-example in a batch, one way to accommodate the noise variance calibrated to the $l_2$ norm of the gradient is to use the max $l_2$ norm measured on per-example gradients in a batch $B$ as the sensitivity of the training function $f_t$ for iteration $t$ (recall Definition~2). 
Consider two scenarios: (1) When any of the per-example gradients in a batch is larger than the clipping bound $C$, the sensitivity is set to $C$, and however, (2) when $l_2$ norm of all per-example gradients in a batch is smaller than the pre-defined clipping bound $C$, the clipping bound $C$ is unfortunately a loose estimation of the true sensitivity of function $f_t$ at iteration $t$. If we instead define the sensitivity of $f_t$ by the max $l_2$ norm among these per-example gradients in the batch, we will correct the problems in the above scenario (2). In summary, the $l_2$-max sensitivity will take the smallest of the max $l_2$ norm and the clipping bound $C$. Given that the $l_2$-norm of the gradients closely follows the trend of gradient changes as the training progresses in iterations,  our dynamic $l_2$ sensitivity approach will have adaptive sensitivity $S$, and hence adaptive noise variance when a fixed $\sigma$ is used, and consequently inject larger noise at the early stage of training and smaller differential privacy noise at the later stage of training. 

Note that unlike mean estimation~\cite{nissim2007smooth,bun2019average}, knowing the $l_2$ norm-based sensitivity of the multi-dimension gradient vector on per-example gradients in deep learning training cannot help disclosing the gradient value at each coordinate. Meanwhile, the $l_2$-max norm actually tracks the sensitivity of the training function at each iteration. Yet composition theorem only requires the individual training function component to be differentially private, with each training function component taking care of its own sensitivity. This makes the $l_2$-max sensitivity a possible choice for sensitivity optimization.

{\bf DP-dynS[C$_{decay}$]. \/} The second dynamic sensitivity strategy is to use dynamic decaying clipping. Given that clipping is used to regulate the largest change of gradients during the training, as the training progresses, the maximum changes of the gradients decrease, and gradually approach zero when the training starts to converge.  Hence, one may use dynamic decaying clipping method to estimate the maximum changes of the gradients and define the dynamic sensitivity accordingly. Motivated by the dynamic learning rate~\cite{darken1990note}, 
we propose a set of decaying policies $DECAY_C(C_0,\gamma, t)$ for implementing the decay-clipping based dynamic sensitivity with $\gamma_1, \gamma_2, \gamma_3$, and $\gamma_4$ being the control parameters for the decaying rate.


\textit{Linear decay}: The clipping decays in a linear trend, defined as $C_t=C_0(1-\gamma_1 t)$ where $\gamma_1>0$ is the smooth controlling term for clipping at iteration $t$.

\textit{Exponential decay}: The clipping decays in an exponential trend defined by an exponential function: $C_t=C_0 e^{-\gamma_2 t}$ where $\gamma_2$ is the control term for exponential trending.


\textit{Cyclic decay}: The clipping decay follows a decaying triangular cyclic policy originated from the cyclic learning rate~\cite{smith2017cyclical}. With step size $V$, $C_t=C_0^* (1-\gamma_3 t)$ for odd $V$ and  $C_t=C_0^* (1+\gamma_3 t)$ for even $V$. $C_0^*$ itself decays every cyclic triangle: $C_0^*=C_0(1-2*\gamma_4 V)$.


\textbf{DP-dynS.\/} The third dynamic sensitivity strategy is to combine $l_2$-max sensitivity and clipping decay sensitivity $C_{decay}$ such that we can take the advantage of the best side of both worlds. Concretely, although $l_2$-max sensitivity is a tighter estimation of the maximum changes in gradients, it may still benefit from additional improvements in some situations, in which the $l_2$-max sensitivity happens to be defined by the clipping bound $C$. In such situations, by integrating the decay clipping sensitivity with the $l_2$-max sensitivity, the decaying clipping $C$ will be used instead of the initial large clipping bound. 
By Lemma~\ref{lemma:gaussian_mechanism}, both the decay clipping-based dynamic sensitivity and $l_2$-max sensitivity do not have a direct impact on the differential privacy guarantee. However, with an accuracy target, the model with combined sensitivity optimizations may reach the accuracy and terminate the training earlier, resulting in smaller accumulated privacy spending. 

\begin{algorithm}[t]
\footnotesize
\caption{\footnotesize DP-dyn[S,$\sigma$]}\label{centralize_dpsgd_12252020-dyn}
\KwIn{ input data $D$, batch size $B$, learning rate $\eta$, maximum iteration $T$, decaying trend $\gamma_C$ and $\gamma_{\sigma}$, loss function: $\mathcal{L}$}
\textbf{initialization:} model $W(0)$, noise scale $\sigma_0$, clipping bound $C_0$.   \\ 
\For{iteration $t$ in $\{0,1,2\dots, T-1\}$}{
 \nonl  \textbf{// batch processing, sampled over $D$ , start $f_t$.} \\
  \nonl  \textbf{// determine the clipping bound for iteration $t$.} \\
  \textcolor{blue}{$C_t=DECAY_C(C_0,\gamma_C, t)$} \\
    \textcolor{blue}{$\sigma_t=DECAY_{\sigma}(\sigma_0,\gamma_{\sigma}, t)$} \\
      \For{instance $i$ in $\{1,...B\}$ }{ 
      \nonl   \textbf{// obtain the per-example gradients for sample $i$ at iteration $t$.}  \\
       $\nabla W(t)_{i} \leftarrow \frac{1}{B} \nabla \mathcal{L}(W(t),i)$ \\
        \nonl   \textbf{// obtain the clipped per-example gradients for sample $i$ at iteration $t$.}  \\
         $\overline \nabla W(t)_i \leftarrow CLIP(\textcolor{blue}{C_t},\nabla W(t)_{i},\phi)$ \\
    }
\textbf{// compute batch gradients on $M$ layers for iteration $t$:} \\
         $ \overline \nabla W(t) \leftarrow \frac{1}{B}\sum \nolimits_{i=1}^B \overline \nabla W(t)_i$ \\  
   \nonl        \textbf{// compute the max of $l_2$ norm over $M$ layers on the batch gradient for iteration $t$, assign the sensitivity $S$. } \\
        \textcolor{blue}{$S \leftarrow \max_i ||\overline \nabla  W(t)_{i}||_2$} \\ 
\nonl   \textbf{// compute sanitized batch gradients.} \\
         $  \widetilde \nabla W(t) \leftarrow \overline \nabla W(t) + \mathcal{N}(0, \textcolor{blue}{\sigma_t^2 S^2}))$ \\  
  \nonl \textbf{// gradient descent.} \\
     $W(t+1) \leftarrow W(t) - \eta \widetilde \nabla W(t)$  \\
        \nonl \textbf{// end $f_{t}$, and start next iteration until reaching $T$.} \\
 }
\textbf{Output:} trained differential private model $W(T)$
\end{algorithm}

\begin{table}[t]
\centering
\scalebox{0.80}{
\small{
\begin{tabular}{|c|c|c|c|c|c|c|}
\hline
           & baseline     & dynS{[}C$_{decay}${]} & dynS{[}$l_2$-max{]} & dynS    & dyn$\sigma$ & dyn{[}S,$\sigma${]} \\ \hline
$C_t$      & $C_0$      & decay               & $C_0$                  & decay      & $C_0$          & decay                  \\ \hline
$S$        & $C_t$      & $C_t$               & $l_2$-max              & $l_2$-max  & $C_t$          & $l_2$-max              \\ \hline
$\sigma_t$ & $\sigma_0$ & $\sigma_0$          & $\sigma_0$             & $\sigma_0$ & decay          & decay                  \\ \hline
\end{tabular}
}}
\caption{\small Comparison of privacy parameters in DP-baseline and the five dynamic privacy parameter approaches}
\label{table:dynamic_summary}
\vspace{-0.2cm}
\end{table}

\subsection{Dynamic Sensitivity with Dynamic $\sigma$}

Dynamic noise scale with a decaying policy is an alternative approach to supporting dynamic Gaussian noise variance over the number of $T$ training iterations, denoted by DP-dyn$\sigma$. Recall that the per-example gradients in early iterations are more informative and thus more vulnerable against gradient leakage attacks. With a dynamic decaying noise scale, we can ensure a larger noise scale and thus larger noise variance in the early stage of training and smoothly decay the scale $\sigma$ as training progresses, such that the per-example gradients are perturbed with larger noise in early iterations and relatively less exploitable by the attacker, even when a fixed sensitivity defined by the fixed clipping bound $C$ is used.
In our prototype implementation, we implement the dynamic decaying noise scale policies with a starting noise scale $\sigma_0$, by employing the same set of three decay policies, denoted by $DECAY_{\sigma}(\sigma_0,\gamma, t)$, used in designing our decay-clipping dynamic sensitivity.

When combining dynamic $\sigma$ with dynamic sensitivity based on both $l_2$-max and decay clipping, denoted by DP-dyn[S,$\sigma$], we expect the best dynamic parameter optimization for differentially private deep learning. 
Algorithm~\ref{centralize_dpsgd_12252020-dyn} provides a sketch of the pseudo-code of DP-dyn[S,$\sigma$], which modifies DP-baseline by adding dynamic decay clipping (line 3), dynamic decay noise scale (line 4), and combining with dynamic $l_2$-max sensitivity (line 11) for each iteration of the training. \textbf{Table~\ref{table:dynamic_summary}} provides a summary of comparison for the five different dynamic privacy parameter optimizations and the DP-baseline with respect to the three privacy parameters: clipping $C$, sensitivity $S$ and noise scale $\sigma$. \textbf{Figure 4} measures the noise variance $\varsigma = S\sigma$ on MNIST with three different settings: (i) fixed $S=C=4$ and fixed $\sigma=6$ (DP-baseline), (ii) $l_2$-max sensitivity and fixed $\sigma$ (DP-dynS) and (iii) DP-dyn[$S$, $\sigma$] with $l_2$-max sensitivity and decaying noise scale starting from $\sigma_0=15$. Compared to the baseline approach with fixed privacy parameters and our methods with dynamic $l_2$-max or dynamic clipping, the method of DP-dyn[$S$, $\sigma$] with both dynamic sensitivity S and dynamic noise scale $\sigma$ offers more resilience and better accuracy because  it can add larger noise in the early stage of training and small noise as the training progresses towards convergence. 



\begin{figure}[t]
 \centerline{\includegraphics[scale=.60]{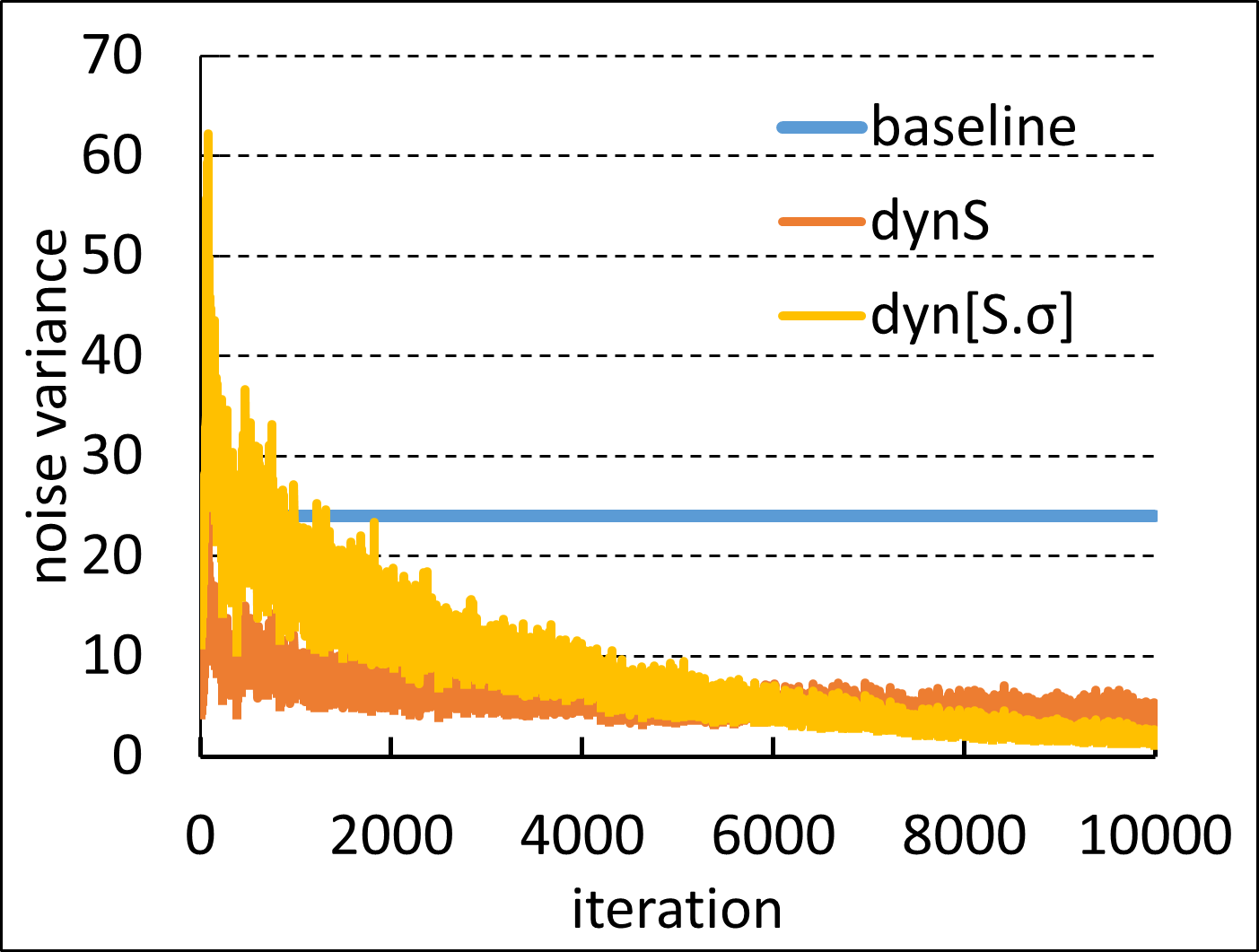}}
 \vspace{-0.1cm}
\captionof{figure}{\small  Noise variance $\varsigma$ of MNIST training under DP-baseline, DP-dynS and DP-dyn[S,$\sigma$].}
 \label{fig:different_decay}
 \vspace{-0.4cm}
\end{figure}

\section{Privacy Analysis Metrics}
\label{metrics}

For DP algorithms with fixed-parameter strategies, such as fixed clipping $C=4$, fixed noise scale $\sigma=6$ and fixed sensitivity $S=C$, no matter which one of the five privacy accounting methods outlined in Section~\ref{sec:preliminary} we choose to use, the total privacy spending $\epsilon$ will be the same, given the sampling rate and $\delta=1e-5$. Hence, the evaluation will focus on measuring and comparing the accuracy performance of the model trained by using these DP-algorithms. However, for DP-algorithms with dynamic privacy parameter strategies, additional metrics should be considered. In this section, we describe four complementary metrics to evaluate and compare the effectiveness of alternative approaches to different differentially private deep learning: (1) model accuracy, (2) differential privacy, (3) resilience against gradient leakage attacks.

\textbf{Accuracy with a target privacy budget.\/} This metric is designed for performing utility analysis with respect to model accuracy under the same differential privacy spending (budget). When the privacy budget $\epsilon$ is given and fixed for comparing different algorithms, according to the privacy accounting methods outlined in Section~\ref{sec:preliminary}, the following parameters are fixed: the total iterations ($T$), the sampling rate $q$, the privacy parameter $\delta$. By Lemma~\ref{lemma:gaussian_mechanism}, the lower bound of noise scale $\sigma$ are also fixed. By Theorem~\ref{theorem:gaussian}, the definition of sensitivity $S$ will directly impact on how the noise variance $\varsigma$ is defined. This motivates us to advocate this metric as one of the four principled criteria to compare the baseline DP algorithm that use fixed privacy parameter strategies (see Section~\ref{sec:dpdl}) with the alternative DP algorithms that use different dynamic privacy parameter strategies (see Section~\ref{sec:adaptive}). 
Following a similar setting, one may conduct privacy analysis to compare alternative DP algorithms under a fixed privacy budget $\epsilon$ and a pre-defined total training iteration ($T$) on different datasets. This allows us to measure the level of differential privacy spending using all five representative privacy accounting methods (recall Section~\ref{sec:preliminary}), and to gain a deeper understanding of the difference among different accounting methods in tracking privacy spending over the same $T$ iterations of training on the same training set. The DP-algorithm with the highest accuracy performance will be the winner in this analysis.  

Next, we introduce two additional privacy analysis metrics: (i) privacy with a target accuracy and a fixed noise scale $\sigma$ and (ii) privacy with a target accuracy and a fixed noise variance $\varsigma$. They are designed specifically to evaluate and compare alternative DP algorithms that use dynamic parameter strategies with baseline DP-algorithms that use fixed-parameter strategies. Both metrics are designed for conducting privacy analysis under the same utility goal defined by the target accuracy. A proper target accuracy should be chosen such that it can be achieved by every alternative algorithm being compared during privacy analysis. Given a fixed target model accuracy, with a fixed sampling rate $q$ and fixed $\delta=1e-5$, some algorithms may terminate the training before reaching its pre-defined total iterations $T$. This may result in spending less privacy budget $\epsilon$, regardless of which one of the five accounting methods is used to track the accumulative privacy spending for privacy analysis. 

\textbf{Privacy with a target accuracy and a fixed $\sigma$.\/} By Lemma~\ref{lemma:gaussian_mechanism}, noise scale $\sigma$ defines the lower bound of the accumulative privacy spending $\epsilon$ over the training iterations used to achieve the given target accuracy. Hence, those DP algorithms that can achieve the target model accuracy before reaching the pre-defined $T$ iterations will terminate the training earlier, which may result in smaller accumulated privacy spending. 
Based on Gaussian Mechanism, when a DP algorithm uses a dynamic sensitivity $S$ with a fixed noise scale $\sigma$, it will result in a dynamically changing noise variance $\varsigma$, following the trend of dynamic sensitivity. In this case, even the fixed $\sigma$ will lead to the same privacy spending $\epsilon$ under the same $T$ iterations, 
the dynamic sensitivity optimized DP training may achieve the target accuracy and terminate earlier. 
This will result in smaller accumulated privacy spending. However, for DP algorithms with fixed privacy parameters, e.g., a fixed noise scale $\sigma$ is combined with a fixed sensitivity $S$, then a constant noise variance$\varsigma$ will be used in each iteration of the training, which may result in excessive noise in the later stage of the training, degrading the utility without any gain on privacy.

\textbf{Privacy with a target accuracy and a fixed $\varsigma$.\/} This metric allows us to conduct privacy analysis from a very different perspective. Based on Theorem~1 and Lemma~1, the Gaussian noise variance is defined as the multiplication of sensitivity $S$ and noise scale $\sigma$. With a fixed noise variance $\varsigma$, all DP-algorithms will inject the same amount of noise in each of the total $T$ training iterations, regardless of whether they are using fixed-parameter strategies or dynamic parameter strategies. However, the accumulated privacy spending for DP-algorithms with dynamic parameter strategies will be smaller compared to the accumulative privacy cost for DP-algorithms with fixed-parameter strategies. Concretely, for DP algorithms with dynamic sensitivity, under a fixed noise variance $\varsigma$, the noise scale $\sigma$ can no longer be fixed, and it is determined in a reverse trend of the dynamic sensitivity $S$. As the training progresses, the dynamic sensitivity tends to align closely with the declining trend of gradient updates, and we will have to use a large noise scale to keep noise variance constant. As a result, DP-algorithms with dynamic sensitivity tend to result in smaller privacy spending ($\epsilon$) compared to the DP-algorithms with fixed parameters.

\textbf{Resilience against Gradient Leakage Attacks.\/} This metric is designed to measure and compare alternative DP-algorithms with respect to attack resilience, which can be defined using both (i) adverse effect of the attack, measured by the accuracy performance under attack, and (ii) attack cost in terms of the time spent to perform reconstruction inference on the leaked gradients and whether the inference results in a successful disclosure of a private training example. A DP-algorithm is considered highly resilient in the effectiveness evaluation, if under the same privacy spending, this algorithm outperforms other alternatives with the highest attack resilience.


\section{Experimental Evaluation}
\label{sec:experiment_utilityprivacy}

\begin{table}[t]
\centering
\scalebox{0.75}{
\small{
\begin{tabular}{|c|c|c|c|c|c|c|c|}
\hline
& \footnotesize{ \# train} &  \footnotesize{\#  test} &  \footnotesize{\#   features} &  \footnotesize{\#   class} &  \footnotesize{\# iter.} &  \footnotesize{batch size} &  \footnotesize{acc.} \\ \hline
 \footnotesize{MNIST}         & 60000              & 10000          & 28*28         & 10           & 10000           & 600          &  \textcolor{red}{0.989}            \\ \hline
 \footnotesize{Fashion-MNIST} & 60000              & 10000          & 28*28         & 10           & 10000           & 600          & \textcolor{red}{0.875}            \\ \hline 
 \footnotesize{CIFAR10}      & 50000              & 10000          & 32*32*3       & 10           & 10000           & 500          & \textcolor{red}{0.687}             \\ \hline
 \footnotesize{LFW}           & 2267               & 756            & 32*32*3       & 62           & 6000            & 22           & \textcolor{red}{0.766}              \\ \hline
 \footnotesize{Purchase-10}   & 10000              & 2000           & 600           & 10           & 5000            & 100          & \textcolor{red}{0.825}            \\ \hline
 \footnotesize{Purchase-50}   & 10000              & 2000           & 600           & 50           & 5000            & 100          & \textcolor{red}{0.667}             \\ \hline
 \end{tabular}
 }}
 \caption{\small Benchmark datasets and parameters}
\label{table:dataset_setup}
\vspace{-0.4cm} 
\end{table}

We evaluate alternative approaches to deep learning with differential privacy using six benchmark datasets. \textbf{Table~\ref{table:dataset_setup}} list the six datasets with training set, test set, \#features, \#classes,  total training iterations $T$, batch size $B$, and non-private accuracy. {\bf MNIST\/} is a grey-scale hand-written digit image dataset. {\bf Fashion-MNIST\/}  is an image dataset associated with a label from 10 clothing classes such as T-shirt, Trouser, and Pullover.  {\bf CIFAR10\/} is a dataset of color images of objects. {\bf LFW\/} is a dataset, originated from 13233 images of 5749 classes, by resizing the original image size of $250\times 250$ to $32\times 32$ and extracting the 'interesting' region. Since most of the classes have a very limited number of data points, we consider 3023 images from 62 classes that have more than 20 images per class. A 4:1 train-test ratio is applied.  {\bf Purchase\/} is a dataset adopted from Kaggle challenge of ``acquire valued shopper", with shopping histories for several thousand individuals. 
We consider a simplified version with 197,325 records, each with 600 binary features converted from the non-numerical raw data. Similar to~\cite{shokri2017membership}, we cluster the data record into 10 and 50 classes for classification. A 4:1 ratio is used for training and test data split. For the four image datasets, a deep convolutional neural network with two convolutional layers and one fully connected layer is used. For the two attribute datasets, a fully connected model with two hidden layers is applied.

\begin{table}[t]
\centering
\scalebox{0.75}{
\small{
\begin{tabular}{|c|c|c|c|c|c|}
\hline
\multicolumn{2}{|c|}{}                                  & mnist    & Fashion-MNIST & cifar10  & LFW      \\ \hline
\multirow{3}{*}{non-private}              & attack iter & 11.5     & 12.4          & 28.3     & 25       \\ \cline{2-6} 
                                          & ASR         &   \textcolor{red}{1}        &  \textcolor{red}{1}             &  \textcolor{red}{0.973}    &  \textcolor{red}{1}        \\ \cline{2-6} 
                                          & MSE         & 1.50E-05 & 2.59E-05      & 2.20E-04 & 2.20E-04 \\ \hline
\multirow{3}{*}{DP-baseline}                   & attack iter & 13.4     & 14.1          & 30.5     & 26.3     \\ \cline{2-6} 
                                          & ASR         & \textcolor{red}{0.23}     & \textcolor{red}{0.25}          & \textcolor{red}{0.16}     & \textcolor{red}{0.13}     \\ \cline{2-6} 
                                          & MSE         & 0.547    & 0.588         & 0.401    & 0.352    \\ \hline
\multirow{3}{*}{Quantileclip~\cite{thakkar2019differentially}}             & attack iter & 12.9     & 14.3          & 30.5     & 27.1     \\ \cline{2-6} 
                                          & ASR         &  \textcolor{red}{0.64}       & \textcolor{red}{0.67}             &  \textcolor{red}{0.47}       &    \textcolor{red}{0.49}     \\ \cline{2-6} 
                                          & MSE         & 0.097    & 0.121         & 0.203    & 0.229    \\ \hline
\multirow{3}{*}{Adaclip~\cite{pichapati2019adaclip}}                  & attack iter & 12.6     & 14.1          & 29.9     & 26.4     \\ \cline{2-6} 
                                          & ASR         & \textcolor{red}{0.72}        & \textcolor{red}{0.76}             & \textcolor{red}{0.58}        & \textcolor{red}{0.61}        \\ \cline{2-6} 
                                          & MSE         & 0.033    & 0.041         & 0.129    & 0.144    \\ \hline
\multirow{3}{*}{DP-dyn{[}$S$,$\sigma${]}} & attack iter     & 300      & 300           & 300      & 300      \\ \cline{2-6} 
                                          & ASR         & \textbf{0}        & \textbf{0}             & \textbf{0}        & \textbf{0}        \\ \cline{2-6} 
                                          & MSE         & 0.886    & 0.904         & 0.912    & 0.897    \\ \hline
\end{tabular}
}}
\caption{{\small Comparing gradient leakage resilience of non-private training with DP-baseline and DP-dyn[S,$\sigma$]. The maximum attack iteration is set to 300.}}
\label{table:leakage_evaluation}
\vspace{-0.6cm}
\end{table}
 \begin{figure}[t]
\centerline{\includegraphics[scale=.63]{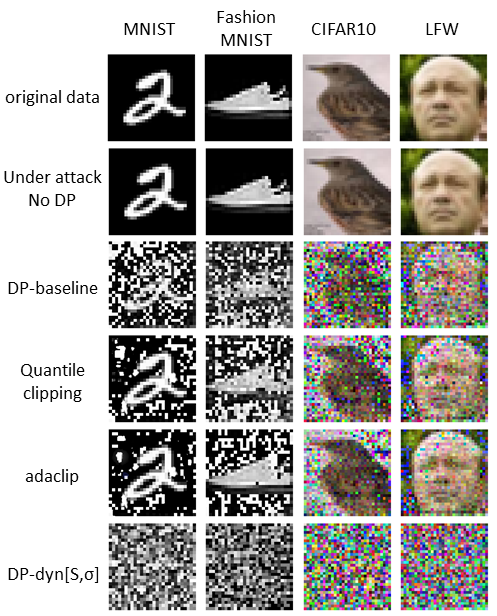}}
\vspace{-0.2cm}
\caption{\small Gradient leakage attack evaluation. Dynamic privacy parameters can help improving the robustness.}
\label{fig:gradient_leakage}
\vspace{-0.4cm}
\end{figure}

\subsection{Resilience against Gradient Leakages}
\label{sec:experiment_robustness}
The first set of experiments measures and compares the resilience of alternative DP algorithms with fixed-parameter strategies, existing adaptive clipping approaches, and dynamic parameter strategies against gradient leakage attacks. 
The following three attack metrics are used to evaluate the adverse effect and cost of gradient leakage attacks, which in turn can be used to measure the resilience of different DP approaches in the presence of attacks. They are attack success rate (ASR), \#attack iterations (attack iter) to succeed the inference with 300 as the default, and attack reconstruction distance in MSE (mean square error). The successful attack reconstruction is defined by MSE smaller than 0.70. We report the attack iterations and attack reconstruction distance only for those successful attack examples. For attacks with complete failure, we record their reconstruction distance at the pre-defined maximum reconstruction/inference iterations of 300. \textbf{Table~\ref{table:leakage_evaluation}} reports the attack results under non-private models, the DP-baseline, DP-dyn[S,$\sigma$], and two representative adaptive clipping methods: quantile-based clipping~\cite{thakkar2019differentially}, Adaclip~\cite{pichapati2019adaclip} and for the four image datasets. 
AdaClip performs the clipping bound estimation based on the coordinates and adaptively add different noise levels to different dimensions of the gradients, whereas the quantile clipping estimates the clipping bound using the quantile of the unclipped gradient norm. 
The default setting of $C=4$, $\sigma=6$ is used for DP-baseline algorithms (fixed parameters) and also used as the initial value for quantile-based clipping~\cite{thakkar2019differentially} and DP-dyn[S,$\sigma$]. The attack is performed on the gradients from the first training iteration as gradients at early training iterations tend to leak more information than gradients in the later stage of the training~\cite{wei2020framework}. \textbf{Figure~\ref{fig:gradient_leakage}} illustrates the reconstruction results. We make three observations: 
(1) For all four datasets, the gradients in non-private models are vulnerable to the gradient leakage attack. 
(2) The quantile-based clipping~\cite{thakkar2019differentially} and Adaclip~\cite{pichapati2019adaclip} method can slightly improve the resilience against the gradient leakage attack compared to non-private training.
 The two adaptive clipping approaches focus mainly on reducing the noise variance rather than gradient attack resilience. Therefore, the DP-baseline, 
 with large and constant noise, is more effective with high resilience against the gradient leakage threats.
(3) Compared to both DP-baseline and existing adaptive clipping approaches, 
DP-dyn[S,$\sigma$] offers the highest resilience against gradient leakage attacks. This is because DP-dyn[S,$\sigma$] adds larger noise at the early iterations of the training thanks to the combined effect of dynamic sensitivity and dynamic noise scale, which effectively creates difficulty for the gradient leakage attack to succeed. The high attack reconstruction distance in Table~\ref{table:leakage_evaluation} shows strong attack resilience against gradient leakage threats. The visualization in Figure~\ref{fig:gradient_leakage} also provides an intuitive illustration of this effect. For the two existing approaches, the adaptive clipping brings down the differential privacy noise compared to DP-baseline and thus shows less improvement on the resilience of DP training against gradient leakage attacks, compared to DP-baseline. However, differential privacy noise injection combined with dynamic sensitivity and dynamic noise scale will offer the best defense against gradient leakage threats.

In the rest of the experiments, we provide empirical results and analysis to further illustrate the benefits of the proposed gradient leakage resilient deep learning with dynamic differential privacy parameter optimizations. We first demonstrate how the proposed DP-dyn{[}$S$,$\sigma${]} achieves higher accuracy compared to the DP-baseline at the same ($\epsilon,\delta$) differential privacy level. Then we show how DP-dyn{[}$S$,$\sigma${]} obtains a stronger differential privacy guarantee in terms of smaller $\epsilon$ spending at a target accuracy. Finally, we show that in addition to high resilience against gradient leakage, the proposed dynamic privacy parameter optimizations have an acceptable time cost compared to the conventional DP deep learning baseline. Furthermore, our approach consistently offers high test accuracy on all benchmark datasets, compared to both the conventional DP deep learning baseline and existing adaptive clipping proposals.

\begin{table}[t]
\centering
\scalebox{0.80}{
\small{
\begin{tabular}{|c|c|c|c|c|c|c|c|}
\hline
\multicolumn{2}{|c|}{}                              & C=1    & C=2    & C=4    & C=8    & C=16   & C=32   \\ \hline
\multirow{5}{*}{\rotatebox{90}{MNIST}}         & non-private        & \multicolumn{6}{c|}{\textcolor{red}{0.9892}}                         \\ \cline{2-8} 
                               & DP-baseline             & 0.9450 & \textcolor{red}{0.963} & 0.960 & 0.945  & 0.915 & 0.881 \\ \cline{2-8} 
                               & DP-dynS[C$_{decay}$]  & 0.948 & 0.966 & 0.964 & 0.951 & 0.931 & 0.919 \\ \cline{2-8} 
                               & DP-dynS[$l_2$-max]     & \textbf{0.959} & 0.975 & 0.977 & 0.983 & 0.975 & 0.976 \\ \cline{2-8} 
                               & DP-dynS & 0.955 & \textbf{0.975} & \textbf{0.978} & \textbf{0.980} & \textbf{0.978} & \textbf{0.977} \\ \hline
\multirow{5}{*}{\rotatebox{90}{\scriptsize Fashion-MNIST}} & non-private        & \multicolumn{6}{c|}{\textcolor{red}{0.875}}                         \\ \cline{2-8} 
                               & DP-baseline             & 0.825  & 0.832 & \textcolor{red}{0.833} & 0.831 & 0.812 & 0.765 \\ \cline{2-8} 
                               & DP-dynS[C$_{decay}$]  & 0.824 & 0.834 & 0.839 & 0.837 & 0.829 & 0.822 \\ \cline{2-8} 
                               & DP-dynS[$l_2$-max]     & \textbf{0.830} & 0.840 & 0.845  & 0.844 & 0.840 & 0.840 \\ \cline{2-8} 
                               & DP-dynS & 0.829 & \textbf{0.841} & \textbf{0.848} & \textbf{0.845} & \textbf{0.840} & \textbf{0.840} \\ \hline
\multirow{5}{*}{\rotatebox{90}{CIFAR10}}      & non-private        & \multicolumn{6}{c|}{\textcolor{red}{0.687}}                          \\ \cline{2-8} 
                               & DP-baseline             & 0.576 & 0.595  & \textcolor{red}{0.608}  & 0.588  & 0.538  & 0.397  \\ \cline{2-8} 
                               & DP-dynS[C$_{decay}$]  & 0.575 & 0.596 & 0.610 & 0.592 & 0.564  & 0.507  \\ \cline{2-8} 
                               & DP-dynS[$l_2$-max]     & \textbf{0.583} & 0.603  & 0.616  & 0.609  & 0.591  & 0.592  \\ \cline{2-8} 
                               & DP-dynS & 0.582  & \textbf{0.604}  & \textbf{0.621}  & \textbf{0.612}  & \textbf{0.595}  & \textbf{0.595}  \\ \hline
\multirow{5}{*}{\rotatebox{90}{LFW}}           & non-private        & \multicolumn{6}{c|}{\textcolor{red}{0.766}}                          \\ \cline{2-8} 
                               & DP-baseline             & 0.704  & \textcolor{red}{0.709}  & 0.692  & 0.647  & 0.595  & 0.406  \\ \cline{2-8} 
                               & DP-dynS[C$_{decay}$]  & 0.692  & 0.714  & 0.703  & 0.675  & 0.643  & 0.606  \\ \cline{2-8} 
                               & DP-dynS[$l_2$-max]     & \textbf{0.715}  & 0.722  & 0.738  & 0.71   & 0.683  & 0.645  \\ \cline{2-8} 
                               & DP-dynS & 0.711  & \textbf{0.725}  & \textbf{0.749}  & \textbf{0.716}  & \textbf{0.685}  & \textbf{0.662}  \\ \hline
\multirow{5}{*}{\rotatebox{90}{Purchase-10}}   & non-private        & \multicolumn{6}{c|}{\textcolor{red}{0.825}}                          \\ \cline{2-8} 
                               & DP-baseline             & 0.731  & 0.734  & \textcolor{red}{0.744}  & 0.739  & 0.716  & 0.683  \\ \cline{2-8} 
                               & DP-dynS[C$_{decay}$]  & 0.727  & 0.738  & 0.749  & 0.744  & 0.729  & 0.721  \\ \cline{2-8} 
                               & DP-dynS[$l_2$-max]     & 0.751  & 0.766  & 0.769  & 0.763  & 0.759  & 0.761  \\ \cline{2-8} 
                               & DP-dynS & \textbf{0.752}  & \textbf{0.768}  & \textbf{0.773}  & \textbf{0.77}   & \textbf{0.765}  & \textbf{0.766}  \\ \hline
\multirow{5}{*}{\rotatebox{90}{Purchase-50}}   & non-private        & \multicolumn{6}{c|}{\textcolor{red}{0.667}}                          \\ \cline{2-8} 
                               & DP-baseline             & 0.549  & 0.561  & 0.571  & \textcolor{red}{0.574}  & 0.556  & 0.503  \\ \cline{2-8} 
                               & DP-dynS[C$_{decay}$]  & 0.545  & 0.569  & 0.577  & 0.581  & 0.579  & 0.562  \\ \cline{2-8} 
                               & DP-dynS[$l_2$-max]     & 0.571  & 0.573  & 0.591  & 0.595  & 0.587  & 0.58   \\ \cline{2-8} 
                               & DP-dynS & \textbf{0.573}  & \textbf{0.577}  & \textbf{0.596}  & \textbf{0.604 } & \textbf{0.591}  & \textbf{0.593}  \\ \hline
\end{tabular}
}}
\caption{\small Effectiveness of dynamic sensitivity optimizations in model accuracy improvement. $\sigma=6$ and $T$ is given in Table~\ref{table:dataset_setup}.}
\label{table:clipping_l2max}
\vspace{-0.4cm}
\end{table}

\subsection{Dynamic Sensitivity: Accuracy Analysis}

This set of experiments compares the DP-baseline with three dynamic sensitivity methods on six datasets: DP-dynS[$l_2$-max], DP-dynS[C$_{decay}$], and DP-dynS, which denotes the dynamic sensitivity defined by combining $l_2$-max sensitivity with dynamic clipping. Given that both DP-baseline and DP-dynS[$l_2$-max] use a constant clipping method with fixed pre-defined clipping bound, for a fair comparison, we vary the setting of clipping bound from 1 to 32 by the interval of the power of 2 and set $\sigma=6$. \textbf{Table~\ref{table:clipping_l2max}} reports the results. We make three observations. 
{\it First}, 
DP-dynS[$l_2$-max] consistently outperforms DP-baseline under all settings of clipping bounds on all six datasets. This is because by Lemma~\ref{lemma:gaussian_mechanism}, DP-baseline uses constant noise variance, and injects constant noise in each iteration of the training, regardless of the decreasing trend of gradients in the later stage of the training. Also, DP-baseline is highly sensitive to the settings of clipping bound for all six datasets. It results in lower accuracy when the clipping bound is set too large (e.g. C=16, 32) or too small bound (e.g. C=1). In contrast, DP-dynS[$l_2$-max] uses dynamic $l_2$-max sensitivity, which changes from iteration to iteration and closely aligns with the declining trend of gradients throughout the training. 
{\it Second}, 
DP-dynS[$l_2$-max] consistently outperforms DP-dynS[C$_{decay}$], showing that noise variance $\varsigma$ defined by $l_2$-max sensitivity and noise scale $\sigma$ offers tighter alignment than dynamic sensitivity defined by decaying clipping bound $C_{decay}$. A simple policy of linearly decaying from the initial clipping bound $C$ to $C/2$ is used in this experiment. 
{\it Third}, 
DP-dynS takes the best from both $l_2$-max sensitivity and dynamic clipping and consistently outperforms all other alternative approaches on all six datasets. 
Based on our empirical observations, it is hard to provide a scalable and stable performance for deep learning with DP property when using dynamic clipping to approximate the sensitivity S for two reasons. (1) Even with dynamic decaying clipping like the DP-dynS[C$_{decay}$] or AdaClip or Quantile Clip (see Table XI), the dynamic clipping relies on the initial setting of the clipping upper bound and the minimum clipping bound as the training progresses toward convergence to ensure the approximation of the sensitivity is correct. Such max and min clipping bound is to some extent dataset and training task dependent, as shown in Table IV. In comparison, our proposed DP-dynS[$l_2$-max] is a scalable and much stable dynamic DP parameter optimization in terms of attack resilience, accuracy, convergence, and privacy spending.

\begin{table}
\centering
\scalebox{0.88}{
\begin{tabular}{|c|c|c|c|c|c|c|c|}
\hline
\multicolumn{2}{|c|}{\multirow{2}{*}{}}             & \multicolumn{5}{c|}{Privacy   spending ($\epsilon$)}    & \multirow{2}{*}{acc} \\ \cline{3-7}
\multicolumn{2}{|c|}{}                              & BaseC & AdvC & OptC & zCDP  & MA &                           \\ \hline
\multirow{4}{*}{\rotatebox{90}{MNIST}}         & DP-baseline             & 123.354  & 7.450   & 6.740   & 1.159 & 0.823   & 0.960                    \\ \cline{2-8} 
                               & DP-dynS[C$_{decay}$]  & 123.354  & 7.450   & 6.740   & 1.159 & 0.823   & 0.964                    \\ \cline{2-8} 
                               & DP-dynS[$l_2$-max]     & 123.354  & 7.450   & 6.740   & 1.159 & 0.823   & 0.977                    \\ \cline{2-8} 
                               & DP-dynS & 123.354  & 7.450   & 6.740   & 1.159 & 0.823   & \textbf{0.978}                    \\ \hline
\multirow{4}{*}{\rotatebox{90}{\tiny Fashion-MNIST}} & DP-baseline             & 123.354  & 7.450   & 6.740   & 1.159 & 0.823   & 0.833                   \\ \cline{2-8} 
                               & DP-dynS[C$_{decay}$]  & 123.354  & 7.450   & 6.740   & 1.159 & 0.823   & 0.839                    \\ \cline{2-8} 
                               & DP-dynS[$l_2$-max]     & 123.354  & 7.450   & 6.740   & 1.159 & 0.823   & 0.845                     \\ \cline{2-8} 
                               & DP-dynS & 123.354  & 7.450   & 6.740   & 1.159 & 0.823   & \textbf{0.848}                    \\ \hline
\multirow{4}{*}{\rotatebox{90}{CIFAR10}}       & DP-baseline             & 123.354  & 7.450   & 6.740   & 1.159 & 0.823   & 0.608                     \\ \cline{2-8} 
                               & DP-dynS[C$_{decay}$]  & 123.354  & 7.450   & 6.740   & 1.159 & 0.823   & 0.610                 \\ \cline{2-8} 
                               & DP-dynS[$l_2$-max]     & 123.354  & 7.450   & 6.740   & 1.159 & 0.823   & 0.616                     \\ \cline{2-8} 
                               & DP-dynS & 123.354  & 7.450   & 6.740   & 1.159 & 0.823   & \textbf{0.621}                     \\ \hline
\multirow{4}{*}{\rotatebox{90}{LFW}}           & DP-baseline             & 74.024   & 5.503   & 5.037   & 0.893 & 0.636   & 0.692                     \\ \cline{2-8} 
                               & DP-dynS[C$_{decay}$]  & 74.024   & 5.503   & 5.037   & 0.893 & 0.636   & 0.703                     \\ \cline{2-8} 
                               & DP-dynS[$l_2$-max]     & 74.024   & 5.503   & 5.037   & 0.893 & 0.636   & 0.738                     \\ \cline{2-8} 
                               & DP-dynS & 74.024   & 5.503   & 5.037   & 0.893 & 0.636   & \textbf{0.749}                     \\ \hline
\multirow{4}{*}{\rotatebox{90}{\scriptsize purchase-10}}   & DP-baseline             & 61.689   & 4.952   & 4.546   & 0.814 & 0.580   & 0.744                     \\ \cline{2-8} 
                               & DP-dynS[C$_{decay}$]  & 61.689   & 4.952   & 4.546   & 0.814 & 0.580   & 0.749                     \\ \cline{2-8} 
                               & DP-dynS[$l_2$-max]     & 61.689   & 4.952   & 4.546   & 0.814 & 0.580   & 0.769                     \\ \cline{2-8} 
                               & DP-dynS & 61.689   & 4.952   & 4.546   & 0.814 & 0.580   & \textbf{0.773}                     \\ \hline
\multirow{4}{*}{\rotatebox{90}{\scriptsize purchase-50}}   & DP-baseline             & 61.689   & 4.952   & 4.546   & 0.814 & 0.580   & 0.571                     \\ \cline{2-8} 
                               & DP-dynS[C$_{decay}$]  & 61.689   & 4.952   & 4.546   & 0.814 & 0.580   & 0.577                     \\ \cline{2-8} 
                               & DP-dynS[$l_2$-max]     & 61.689   & 4.952   & 4.546   & 0.814 & 0.580   & 0.591                     \\ \cline{2-8} 
                               & DP-dynS & 61.689   & 4.952   & 4.546   & 0.814 & 0.580   & \textbf{0.596}                     \\ \hline
\end{tabular}
}
\captionof{table}{\small Measure and compare privacy spending $\epsilon$ using the five different privacy composition methods for DP-baseline, DP-dynS[$l_2$-max], DP-dynS[C$_{decay}$] and DP-dynS, with clipping bound $C=4$, noise scale $\sigma=6$, and $\delta=1e-5$.}
\label{table:privacy_evaluation} 
\vspace{-0.4cm}
\end{table}

\subsection{Dynamic Sensitivity: Privacy Analysis}
\textbf{Privacy under five accounting methods.\/} The first set of experiments for privacy analysis is designed to measure privacy spending using five privacy accounting methods for DP-baseline and three alternative optimizations of dynamic sensitivity: DP-dynS[$l_2$-max], DP-dynS[C$_{decay}$] and DP-dynS. \textbf{Table~\ref{table:privacy_evaluation}} reports the results. We make two observations. First, all five accounting methods show consistent performance for all four DP algorithms on the six benchmark datasets, with Moment Accountant as the most efficient tracking of privacy spending followed by zCDP. Second, DP-dynS consistently outperforms both DP-baseline and the other two dynamic sensitivity algorithms with high accuracy performance, showing that the combo of $l_2$ sensitivity with decay clipping method $C_{decay}$ is more effective in enabling the sensitivity $S$ to be aligned with the trend of gradient updates during the training.

\textbf{Privacy under a target accuracy and a fixed $\sigma$.\/} In this second set of experiments, we analyze and compare DP-baseline with three alternative dynamic sensitivity optimizations in terms of their privacy spending under a target accuracy and a fixed noise scale $\sigma$. We use the accuracy achieved by DP-baseline at $T$ iterations as the target accuracy for each of the six datasets (recall Table~2). \textbf{Table~\ref{table:targetacc_changingrounds}} reports the results. DP-dynS is the winner consistently across all six datasets because it is the first to achieve the target accuracy with the smallest privacy spending and hence it is the first to terminate the training. Concretely, it takes 5137, 5532, 6336, 3645,  3690, and 3995 iterations for DP-dynS to achieve the target accuracy for MNIST, Fashion-MNIST, CIFAR10, LFW, Purchase-10, and Purchase-50 respectively. As a result, its accumulated privacy spending is much less than the other three approaches.

\begin{table}[t]
\centering
\scalebox{0.70}{
\small{
\begin{tabular}{|c|c|c|c|c|c|c|c|}
\hline
\multicolumn{2}{|c|}{\multirow{2}{*}{}}                                                                & \multicolumn{5}{c|}{Privacy   spending ($\epsilon$)} & \multirow{2}{*}{iter} \\ \cline{3-7}
\multicolumn{2}{|c|}{}                                                                                 & BaseC    & AdvC   & OptC   & zCDP   & MA    &                       \\ \hline
\multirow{4}{*}{\begin{tabular}[c]{@{}c@{}}MNIST \\ (0.960)\end{tabular}}         & DP-baseline             & 123.354  & 7.450  & 6.740  & 1.159  & 0.823 & 10000                 \\ \cline{2-8} 
                                                                               & DP-dynS[C$_{decay}$]  & 98.646   & 6.518  & 5.930  & 1.034  & 0.735 & 7996                  \\ \cline{2-8} 
                                                                                  & DP-dynS[$l_2$-max]     & 66.858   & 5.188  & 4.756  & 0.848  & 0.604 & 5419                  \\ \cline{2-8} 
                                                                                  & DP-dynS & \textbf{63.379}   & \textbf{5.029}  & \textbf{4.615}  & \textbf{0.825}  & \textbf{0.588} & 5137                  \\ \hline
\multirow{4}{*}{\begin{tabular}[c]{@{}c@{}}\footnotesize{Fashion-MNIST} \\ (0.833)\end{tabular}} & DP-baseline             & 123.354  & 7.450  & 6.740  & 1.159  & 0.823 & 10000                 \\ \cline{2-8} 
                                                                                & DP-dynS[C$_{decay}$]  & 100.200  & 6.578  & 5.983  & 1.042  & 0.741 & 8124                  \\ \cline{2-8} 
                                                                                  & DP-dynS[$l_2$-max]     & 71.878   & 5.411  & 4.955  & 0.880  & 0.626 & 5826                  \\ \cline{2-8} 
                                                                                  & DP-dynS & \textbf{68.251}   & \textbf{5.250}  & \textbf{4.812}  & \textbf{0.857}  & \textbf{0.610} & 5532                  \\ \hline
\multirow{4}{*}{\begin{tabular}[c]{@{}c@{}}CIFAR10 \\   (0.608)\end{tabular}}     & DP-baseline             & 123.354  & 7.450  & 6.740  & 1.159  & 0.823 & 10000                 \\ \cline{2-8} 
                                                                                & DP-dynS[$C_{decay}$]  & 109.044  & 6.919  & 6.280  & 1.088  & 0.773 & 8841                  \\ \cline{2-8} 
                                                                                  & DP-dynS[$l_2$-max]     & 80.809   & 5.794  & 5.294  & 0.934  & 0.664 & 6550                  \\ \cline{2-8} 
                                                                                  & DP-dynS & \textbf{78.169}   & \textbf{5.682}  & \textbf{5.195}  & \textbf{0.918}  & \textbf{0.653} & 6336                  \\ \hline
\multirow{4}{*}{\begin{tabular}[c]{@{}c@{}}LFW   \\ (0.692)\end{tabular}}         & DP-baseline             & 74.024   & 5.503  & 5.037  & 0.893  & 0.636 & 6000                  \\ \cline{2-8} 
                                                                                 & DP-dynS[$C_{decay}$]  & 62.417   & 4.985  & 4.576  & 0.819  & 0.583 & 5059                  \\ \cline{2-8}  
                                                                                  & DP-dynS[$l_2$-max]     & 47.331   & 4.254  & 3.919  & 0.711  & 0.508 & 3836                  \\ \cline{2-8} 
                                                                                  & DP-dynS & \textbf{44.975}   & \textbf{4.132}  & \textbf{3.809}  & \textbf{0.693}  & \textbf{0.495} & 3645                  \\ \hline
\multirow{4}{*}{\begin{tabular}[c]{@{}c@{}}purchase-10   \\ (0.744)\end{tabular}} & DP-baseline             & 61.689   & 4.952  & 4.546  & 0.814  & 0.580 & 5000                  \\ \cline{2-8} 
                                                                                & DP-dynS[$C_{decay}$]  & 58.926   & 4.822  & 4.430  & 0.795  & 0.567 & 4776                  \\ \cline{2-8} 
                                                                                  & DP-dynS[$l_2$-max]     & 46.615   & 4.217  & 3.886  & 0.706  & 0.504 & 3778                  \\ \cline{2-8} 
                                                                                  & DP-dynS & \textbf{45.530}   & \textbf{4.161}  & \textbf{3.835}  & \textbf{0.697}  & \textbf{0.498} & 3690                  \\ \hline
\multirow{4}{*}{\begin{tabular}[c]{@{}c@{}}purchase-50 \\ (0.571)\end{tabular}}   & DP-baseline             & 61.689   & 4.952  & 4.546  & 0.814  & 0.580 & 5000                  \\ \cline{2-8} 
                                                                                & DP-dynS[C$_{decay}$]  & 59.308   & 4.840  & 4.446  & 0.798  & 0.568 & 4807                  \\ \cline{2-8} 
                                                                                  & DP-dynS[$l_2$-max]     & 50.871   & 4.433  & 4.080  & 0.738  & 0.526 & 4123                  \\ \cline{2-8} 
                                                                                  & DP-dynS & \textbf{49.292}   & \textbf{4.354}  & \textbf{4.009}  & \textbf{0.726}  & \textbf{0.518} & 3995                  \\ \hline
\end{tabular}
}}
\caption{\small Measure and compare differential privacy spending $\epsilon$ under the target accuracy and fixed noise scale $\sigma=6$ for DP-baseline, DP-dynS[$l_2$-max], DP-dynS[C$_{decay}$] and DP-dynS. Clipping bound $C=4$, $\delta=1e-5$. The target accuracy is set to the accuracy of DP-baseline of each dataset.}
\label{table:targetacc_changingrounds}
\vspace{-0.2cm}
\end{table}

\begin{table}[t]
\centering
\scalebox{0.70}{
\small{
\begin{tabular}{|c|c|c|c|c|c|c|c|}
\hline
\multicolumn{2}{|c|}{}                                                                         & BaseC   & AdvC    & OptC    & zCDP  & MA     & acc    \\ \hline
\multirow{4}{*}{\begin{tabular}[c]{@{}c@{}}MNIST\\ (C=4)\end{tabular}}    & DP-baseline             & 123.354 & 7.450   & 6.740   & 1.159 & 0.823  & 0.9596 \\ \cline{2-8} 
                                                                          & DP-dynS[C$_{decay}$]  & 105.173 & 5.889   & 5.736   & 1.119 & 0.806  & 0.9596 \\ \cline{2-8} 
                                                                          & DP-dynS[$l_2$-max]     & 34.412  & 1.866   & 1.787   & 0.436 & 0.314  & 0.9596 \\ \cline{2-8} 
                                                                          & DP-dynS & \textbf{32.529}  & \textbf{1.783}   & \textbf{1.694}   & \textbf{0.412} & \textbf{0.307}  & 0.9596 \\ \hline
\multirow{4}{*}{\begin{tabular}[c]{@{}c@{}}MNIST\\ (C=8)\end{tabular}}    & DP-baseline             & 123.354 & 7.450   & 6.740   & 1.159 & 0.823  & 0.945  \\ \cline{2-8} 
                                                                          & DP-dynS[C$_{decay}$]  & 114.531 & 6.552   & 6.341   & 0.994 & 0.675  & 0.945  \\ \cline{2-8} 
                                                                          & DP-dynS[$l_2$-max]     & 28.828  & 1.589   & 1.511   & 0.378 & 0.283  & 0.945  \\ \cline{2-8} 
                                                                          & DP-dynS & \textbf{28.015}  & \textbf{1.426}   & \textbf{1.412}   & \textbf{0.366} & \textbf{0.269}  & 0.945  \\ \hline
\multirow{4}{*}{\begin{tabular}[c]{@{}c@{}}CIFAR10 \\ (C=4)\end{tabular}} & DP-baseline             & 123.354 & 7.450   & 6.740   & 1.159 & 0.823  & 0.608  \\ \cline{2-8} 
                                                                          & DP-dynS[C$_{decay}$]  & 120.119 & 7.006   & 6.662   & 1.153 & 0.812  & 0.608  \\ \cline{2-8} 
                                                                          & DP-dynS[$l_2$-max]     & 116.717 & 6.292   & 6.330   & 1.114 & 0.790  & 0.608  \\ \cline{2-8} 
                                                                          & DP-dynS & \textbf{113.458} & \textbf{5.557}   & \textbf{5.536}   & \textbf{1.044} & \textbf{0.763}  & 0.608  \\ \hline
\multirow{4}{*}{\begin{tabular}[c]{@{}c@{}}CIFAR10 \\ (C=8)\end{tabular}} & DP-baseline             & 123.354 & 7.450   & 6.740   & 1.159 & 0.823  & 0.588  \\ \cline{2-8} 
                                                                          & DP-dynS[C$_{decay}$]  & 116.304 & 6.316   & 6.405   & 1.137 & 0.792  & 0.588  \\ \cline{2-8} 
                                                                          & DP-dynS[$l_2$-max]     & 109.405 & 5.865   & 5.886   & 1.063 & 0.754  & 0.588  \\ \cline{2-8} 
                                                                          & DP-dynS & \textbf{105.256} & \textbf{5.849}   & \textbf{5.834}   & \textbf{1.022} & \textbf{0.748}  & 0.588  \\ \hline
 \multirow{4}{*}{\begin{tabular}[c]{@{}c@{}}MNIST\\ (C=4)\end{tabular}}    & DP-baseline             & 523.375 & 232.117 & 258.635 & 4.109 & 21.174 & 0.9773 \\ \cline{2-8} 
                                                                          & DP-dynS[C$_{decay}$]  & 155.677 & 8.575   & 9.037   & 1.370 & 0.944  & 0.9773 \\ \cline{2-8} 
                                                                          & DP-dynS[$l_2$-max]     & 123.354 & 7.450   & 6.740   & 1.159 & 0.823  & 0.9773 \\ \cline{2-8} 
                                                                          & DP-dynS & \textbf{117.266} & \textbf{7.036}   & \textbf{6.687}   & \textbf{1.149} & \textbf{0.806}  & 0.9773 \\ \hline
\end{tabular}
}}
\caption{\small Measure and compare differential privacy spending $\epsilon$ under a target accuracy and a fixed noise variance $\varsigma$ for DP-baseline, DP-dynS[$l_2$-max], DP-dynS[C$_{decay}$] and DP-dynS. Clipping bound $C=4$ or $C=8$, fixed noise scale $\sigma=6$ and $\delta=1e-5$. The target accuracy is set to the accuracy of DP-baseline for both MNIST and CIFAR10 in the first four scenarios. 
For the last scenario, the target accuracy is set to the accuracy of DP-dynS[$l_2$-max] (0.9773) for MNIST with $C=4$.
}
\label{table:targetacc}
\vspace{-0.4cm}
\end{table}

\textbf{Privacy under a target accuracy and a fixed $\varsigma$. \/} 
The third set of experiments for privacy analysis measures the privacy spending under a target accuracy and a fixed noise variance $\varsigma$. 
We compare DP-baseline and the three alternative DP algorithms with three different dynamic sensitivity strategies DP-dynS[$l_2$], DP-dynS[$C_{decay}$], and DP-dynS on MNIST and CIFAR10. The accuracy of DP-baseline is used as the target accuracy for both MNIST and CIFAR10. We set $\sigma=6$, $C=4$ and $C=8$ and conduct the experiments with two fixed  noise variance settings: $\varsigma=24$ with $C=4$ and $\varsigma=48$ with $C=8$. Based on Theorem~2 and Lemma~1, when the noise variance $\varsigma$ is fixed, with dynamic sensitivity $S$, the noise scale $\sigma$ will follow the trend of $S$ in a reverse trend: as $S$ continues to decline with the training progresses in $\#$iterations, the noise scale $\sigma$ will increase to keep the noise scale $\sigma$ constant. As a result, larger sensitivity $S$ in the early stage of the training will lead to a smaller noise scale and larger privacy spending to protect informative gradients. As the training is approaching the end, the sensitivity $S$ will become smaller, resulting in larger $\sigma$ and smaller privacy spending. \textbf{Table~\ref{table:targetacc}} shows the results. Consider the first four scenarios, in which the target accuracy is set to the accuracy of DP-baseline for each dataset. We make three observations: 
(1) DP-dynS consistently outperforms the other three alternatives for the two noise variance settings on both datasets when given a target accuracy and a fixed noise variance $\varsigma$. 
(2) 
DP-dynS[$l_2$-max] consistently ranked as the second winner with significantly smaller privacy spending than DP-dynS[$C_{decay}$] for all four scenarios under all five privacy accounting methods. This further demonstrates that using adaptive clipping with decay function is still a loose estimation compared to using the $l_2$-max sensitivity for differentially private deep learning. Hence, DP-dynS[$C_{decay}$] consumes more privacy budget under both fixed noise variances for both datasets, compared to DP-dynS[$l_{2}$-max] and DP-dynS.
(3) 
When comparing the privacy spending on the same dataset with two different fixed noise variances, or when comparing two different datasets on the same fixed noise variance, it is interesting to note that DP-baseline has the same privacy spending under both variance settings on both MNIST and CIFAR10. This is because the privacy spending $\epsilon$ is anti-correlated to noise scale $\sigma$ (recall Lemma~1) and not sensitive to the different settings of clipping bound for the constant clipping method. However, for DP algorithms with dynamic sensitivity, the privacy spending is also training data dependent. For each of the two fixed noise variances, both with $\sigma=6$, all three dynamic sensitivity algorithms on CIFAR10 will incur higher privacy spending while their respective privacy spending on MNIST will be smaller. This confirms the common knowledge that the gradient update trend during training is model-dependent and dataset-dependent.

The last scenario is included for privacy analysis under two different target accuracy. For MNIST with $C=4$, we set the target accuracy to the accuracy of DP-dynS[$l_2$-max] (0.9773) instead of 0.9596 from DP-baseline. We make two interesting observations. (1) DP-dynS remains the winner with strong privacy guarantee at the smallest privacy spending $\epsilon$, followed by DP-dynS[$C_{decay}$]. (2) To achieve the target accuracy of 0.9773, DP-baseline has to enforce a smaller $\sigma$ to maintain the fixed $\varsigma$, resulting in much higher spending of privacy budget according to all five privacy accounting methods. For MA, DP-baseline results in $\epsilon=21.174$ for target accuracy of 0.9773 compared to $\epsilon=0.823$ for the target accuracy of 0.9596 obtained at $T=10000$. For zCDP and base composition, DP-baseline spent $4\times$ of the privacy budget for achieving the target accuracy of 0.9773, compared to the privacy spent for achieving the target accuracy of 0.9596 at $T=10000$.

\subsection{Dynamic Noise Scale Optimization}
In this section, we evaluate the effectiveness of incorporating dynamic noise scale into DP-baseline and the three alternative dynamic sensitivity optimized DP algorithms.

 \begin{table}[t]
\centering
\scalebox{0.70}{
\small{
\begin{tabular}{|c|c|c|c|c|c|c|}
\hline
\multicolumn{2}{|c|}{\multirow{2}{*}{}}                             & BaseC   & AdvC   & OptC   & zCDP   & MA     \\ \cline{3-7} 
\multicolumn{2}{|c|}{}                                              & 123.354 & 7.450  & 6.740  & 1.159  & 0.823  \\ \hline
\multirow{5}{*}{fix   $\sigma$} & DP-baseline                            & \multicolumn{5}{c|}{0.9596}                  \\ \cline{2-7} 
                                & DP-dynS[C$_{decay}$]                 & \multicolumn{5}{c|}{0.9639}                  \\ \cline{2-7} 
                                & DP-dynS[$l_2$-max]                    & \multicolumn{5}{c|}{0.9773}                  \\ \cline{2-7} 
                                & DP-dynS                & \multicolumn{5}{c|}{\textcolor{red}{0.9778}}                  \\ \cline{2-7} 
                                & iteration                         & \multicolumn{5}{c|}{10000}                  \\ \hline
\multirow{5}{*}{\begin{tabular}[c]{@{}c@{}}adaptive $\sigma$\\ (linear)\end{tabular}}         & DP-dyn$\sigma$                    & 0.962   & 0.9614  & 0.9608 & 0.9618 & 0.9614 \\ \cline{2-7} 
                                & DP-dyn{[}S[C$_{decay}$],$\sigma${]}  & 0.9673  & 0.965  & 0.9641 & 0.9653 & 0.9649 \\ \cline{2-7} 
                                & DP-dyn{[}S[$l_2$-max],$\sigma${]}     & 0.9788  & 0.978  & 0.9775 & 0.9781 & 0.9778 \\ \cline{2-7} 
                                & DP-dyn{[}S,$\sigma${]} & \textbf{0.9791}  & \textbf{0.9783} & \textbf{0.9779} & \textbf{0.9783} & \textbf{0.9782} \\ \cline{2-7} 
                                & Iteration                         & 9646    & 9481   & 8908   & 9494   & 9450   \\ \hline
\multirow{5}{*}{\begin{tabular}[c]{@{}c@{}}adaptive $\sigma$\\ (exponential)\end{tabular}}    & DP-dyn$\sigma$                    & 0.9616  & 0.9604 & 0.9597 & 0.9608 & 0.9603 \\ \cline{2-7} 
                                & DP-dyn{[}S[C$_{decay}$],$\sigma${]}  & 0.9681  & 0.9652 & 0.9647 & 0.9665 & 0.9665 \\ \cline{2-7} 
                                & DP-dyn{[}S[$l_2$-max],$\sigma${]}     & 0.979   & 0.9775 & 0.9775 & 0.9786 & 0.9781 \\ \cline{2-7} 
                                & DP-dyn{[}S,$\sigma${]} & \textbf{0.9803}  & \textbf{0.9781} & \textbf{0.9781} & \textbf{0.9789} & \textbf{0.9787} \\ \cline{2-7} 
                                & iteration                         & 9678    & 9311   & 8955   & 9553   & 9501   \\ \hline
\multirow{5}{*}{\begin{tabular}[c]{@{}c@{}}adaptive $\sigma$\\ (cyclic)\end{tabular}}         & DP-dyn$\sigma$                    & 0.9601  & 0.9599 & 0.9596 & 0.9599 & 0.9599 \\ \cline{2-7} 
                                & DP-dyn{[}S[C$_{decay}$],$\sigma${]}  & 0.9652  & 0.9644 & 0.9641 & 0.9648 & 0.9645 \\ \cline{2-7} 
                                & DP-dyn{[}S[$l_2$-max],$\sigma${]}     & 0.9784  & 0.9779 & 0.9776 & 0.9781 & 0.9779 \\ \cline{2-7} 
                                & DP-dyn{[}S,$\sigma${]} & \textbf{0.9787}  & \textbf{0.9781} & \textbf{0.9779} & \textbf{0.9785} & \textbf{0.9784} \\ \cline{2-7} 
                                & iteration                         & 8019    & 7724   & 7251   & 7993   & 7876   \\ \hline
\end{tabular}
}}
\caption{{\small Accuracy analysis for eight alternative DP algorithms on MNIST under three different dynamic noise scale policies ($C=4$, $\delta=1e-5$). The target $\epsilon$ is from  Table~\ref{table:privacy_evaluation}, $\sigma=6$ for DP algorithms with fixed $\sigma$ and $\sigma=6$ is also used as the initial $\sigma$ value for dynamic $\sigma$ algorithms under each of the three decaying policies.}
}
\label{table:epsguidednoisedecay}
\vspace{-0.2cm}
\end{table} 

{\bf Accuracy analysis under a target privacy budget $\epsilon$.\/} The first set of experiments evaluate and compare these eight alternative DP algorithms on MNIST under a given target privacy budget $\epsilon$ such that each algorithm will terminate when its target privacy budget is exhausted. Unless otherwise stated, $\delta=1e-5$, $C=4$ and $T=10000$ for MNIST. 
\textbf{Table~\ref{table:epsguidednoisedecay}} reports the results. We make three observations. 
(1) For DP-baseline and the three dynamic sensitivity algorithms under fixed $\sigma=6$, we measure and compare their accuracy since they have the same privacy spending as shown in Table~\ref{table:privacy_evaluation} when reaching $T$ iterations. DP-dynS and DP-dynS[$l_2$-max] are clear winners with DP-dynS slightly higher in accuracy.  
(2) By Lemma~1, the decaying noise scale will lead to increased per-iteration $\epsilon$ spending for early iterations in the training. As a result, the training is terminated early when the accumulated privacy loss reaches the target privacy budget under each accounting method. Different privacy composition methods accumulate heterogeneous privacy spending differently and result in different ending iterations. 
(3) By integrating dynamic noise scale optimization, one can further improve accuracy performance under all three noise scale decay policies compared to their corresponding algorithms under the fixed $\sigma$. DP-dyn[S,$\sigma$] is consistently the best performing algorithm under all three $\sigma$ decaying policies. 
Although DP-dyn$\sigma$ with only dynamic noise scale 
slightly outperforms DP-baseline under the three decaying $\sigma$ policies, dynamic sensitivity approaches consistently outperform DP-dyn$\sigma$ in accuracy performance, showing the critical role of sensitivity in achieving high training utility of DP algorithms.  
(4) Empirically, noise scale decay with the exponential trend has the best accuracy performance.

\begin{table}[t]
\centering
\scalebox{0.73}{
\small{
\begin{tabular}{|c|c|c|c|c|c|c|c|}
\hline
\multicolumn{2}{|c|}{}                                                               & BaseC   & AdvC  & OptC  & zCDP  & MA    & iter  \\ \hline
\multirow{4}{*}{fix   $\sigma$} & DP-baseline                                             & 123.354 & 7.450 & 6.740 & 1.159 & 0.823 & 10000 \\ \cline{2-8} 
                                & DP-dynS[C$_{decay}$]                                 & 98.646  & 6.518 & 5.930 & 1.034 & 0.735 & 7996  \\ \cline{2-8} 
                                & DP-dynS[$l_2$-max]                                     & 66.858  & 5.188 & 4.756 & 0.848 & 0.604 & 5419  \\ \cline{2-8} 
                                & DP-dynS                                 & 63.379  & 5.029 & 4.615 & 0.825 & 0.588 & 5137  \\ \hline
linear                          & \multirow{3}{*}{DP-dyn{[}S,$\sigma${]}} & 51.365  & 4.538 & 4.501 & 0.561 & 0.459 & 5532  \\ \cline{1-1} \cline{3-8} 
exponential                     &                                                    & \textbf{48.027}  & \textbf{4.017} & \textbf{3.962} & \textbf{0.537} & \textbf{0.453} & 5541  \\ \cline{1-1} \cline{3-8} 
cyclic                          &                                                    & 56.776  & 4.703 & 4.667 & 0.683 & 0.572 & 5493  \\ \hline
\end{tabular}
}}
\caption{{\small Comparing privacy spending $\epsilon$ for the best dynamic parameter combo DP-dyn[S,$\sigma$] with DP-baseline and three  dynamic sensitivity optimizations under fixed noise scale. The target accuracy is set to 0.9596, the accuracy of DP-baseline on MNIST with $C=4$, $\sigma=6$, $\delta=1e-5$.}
}
\label{table:targetacc_changingrounds_noisescaledecay}
\vspace{-0.4cm}
\end{table}

\textbf{Privacy analysis under a target accuracy.\/} In the next set of experiments, we measure and compare the accumulated privacy spending under a given target accuracy to evaluate the effectiveness of the best dynamic parameter optimization DP-dyn[S,$\sigma$] by comparing it with DP-baseline and three alternative dynamic sensitivity algorithms with fixed $\sigma$ for MNIST: DP-dynS[$l_2$-max], DP-dynS[C$_{decay}$], DP-dynS and DP-dyn{[}S,$\sigma${]}. The target accuracy is set to 0.9596, the accuracy of DP-baseline as provided in Table~\ref{table:privacy_evaluation} with $C=4$ and $\sigma=6$. 
\textbf{Table~\ref{table:targetacc_changingrounds_noisescaledecay}} reports the results. We make two observations: (1) DP-dyn{[}S,$\sigma${]} with exponential $\sigma$ decay provides the best differential privacy guarantee at the smallest privacy spending $\epsilon$ under all five privacy accounting methods. 
(2) Both DP-dynS and DP-dynS[$l_2$-max] terminate slightly earlier than DP-dyn[S,$\sigma$] under the same target accuracy. This is likely because the large early noise in the dynamic noise scale algorithm may have some marginal effect on the convergence, leading to taking a few iterations more than the DP-dynS in reaching the target accuracy. However, DP-dyn[S,$\sigma$] spent the smallest privacy budget to achieve the target accuracy even with a slightly longer training time. This also shows that early termination does not always guarantee smaller privacy spending. The dynamic sensitivity and dynamic noise scale ultimately control the right  amount of noise to be added for achieving the best privacy under a target accuracy.

\subsection{Time Cost Evaluation}
This set of experiments compare the time cost for the non-private algorithm and five alternative approaches to differentially private deep learning
using all six benchmark datasets. \textbf{Table~\ref{table:timecost}} report the per-iteration time cost measurement in seconds. We make three observations: (1) Incorporating dynamic parameters of sensitivity and noise scale incurs negligible additional cost. This is because all differentially private deep learning algorithms will always need to compute the $l_2$ norm of the gradients in order to perform clipping. The difference between DP algorithms and non-private model indicates the cost of computing $l_2$ norm of the gradients per iteration. (2) The relative cost is smaller when the model is simpler with a smaller number of parameters. For example, 
the two attribute datasets have a lower time cost than the four image datasets. (3) In addition to model complexity, batch size may also impact this additional time cost. The relative cost is smaller when the batch size is smaller. For example, LFW has a much smaller batch size than the other three image datasets and is relatively faster to run one iteration.

\begin{table}[t]
\centering
\scalebox{0.77}{
\small{
\begin{tabular}{|c|c|c|c|c|c|}
\hline
                                  & MNIST  & \footnotesize{Fashion-MNIST} & CIFAR10 & LFW    & \footnotesize{Purchase}  \\ \hline
non-private                       & 0.138 & 0.144        & 0.504  & 0.069 & 0.024           \\ \hline
DP-baseline                            & 1.63  & 1.68         & 1.8    & 0.149 & 0.125           \\ \hline
DP-dynS[C$_{decay}$]                 & 1.70  & 1.72         & 1.83   & 0.152 & 0.131           \\ \hline
DP-dynS[$l_2$-max]                    & 1.71  & 1.72         & 1.83   & 0.153 & 0.131           \\ \hline
DP-dynS                & 1.72  & 1.74         & 1.84   & 0.155 & 0.132         \\ \hline
DP-dyn{[}S,$\sigma${]} & 1.72  & 1.74        & 1.85   & 0.155 & 0.132          \\ \hline
\end{tabular}
}}
\caption{\small Per-iteration time cost in seconds}
\label{table:timecost}
\vspace{-0.4cm}
\end{table}

\subsection{Comparison with Existing Dynamic Clipping}

We have shown in Section~\ref{sec:experiment_robustness} that our proposed approach for gradient leakage resilient deep learning with dynamic parameter optimizations offers the strongest resilience against gradient leakage attacks to training data privacy by comparing with both the state of the art DP baseline for deep learning and the two representative adaptive clipping enhancements~\cite{pichapati2019adaclip,thakkar2019differentially}. In this set of experiments, we provide the accuracy performance comparison to show that our dynamic parameter optimization on both noise scale and sensitivity outperforms both the baseline DP approach~\cite{abadi2016deep} and the two adaptive clipping proposals. 
\textbf{Table~\ref{table:adaptive_clipping}} compares DP-dyn{[}S,$\sigma${]} with the two approaches in terms of accuracy performance and time cost. We consider the accuracy for the four image datasets and measure the time cost by sec/iteration. We make two observations: (1) The proposed dynamic optimization on both sensitivity $S$ and noise-scale $\sigma$ consistently outperforms both baseline DP and the two existing proposals to improve the baseline DP with AdaClip or Quantile clipping with larger accuracy performance enhancements. (2) Comparing to the baseline DP~\cite{abadi2016deep}, the proposed dynamic parameter optimization approach incurs the smallest time cost compared to AdaClip and Quantile clipping while offering the highest accuracy improvement in comparison. 

\begin{table}[t]
\centering
\scalebox{0.76}{
\small{
\begin{tabular}{|c|c|c|c|c|c|}
\hline
\multicolumn{2}{|c|}{}                                                                      & MNIST & Fashion-MNIST & CIFAR-10 & LFW   \\ \hline
\multirow{2}{*}{DP-baseline} & accuracy  & 0.960 & 0.833         & 0.608    & 0.692 \\ \cline{2-6} 
                                                                                     & cost  (s/iter) & 1.63  & 1.68          & 1.8     & 0.149 \\ \hline
\multirow{2}{*}{Quantileclip~\cite{thakkar2019differentially}} & accuracy  & 0.971 & 0.846         & 0.614    & 0.733 \\ \cline{2-6} 
                                                                                     & cost (s/iter) & 2.66  & 2.75          & 2.92     & 0.304 \\ \hline
\multirow{2}{*}{Adaclip~\cite{pichapati2019adaclip}}           & accuracy  & 0.969 & 0.843         & 0.611    & 0.725 \\ \cline{2-6} 
                                                                                     & cost (s/iter) & 1.81  & 1.84          & 1.94     & 0.185 \\ \hline
\multirow{2}{*}{DP-dyn{[}S,$\sigma${]}}                                      & accuracy  & \textbf{0.978} & \textbf{0.848}         & \textbf{0.621}    & \textbf{0.749} \\ \cline{2-6} 
                                                                                     & cost (s/iter) & \textbf{1.72}  & \textbf{1.74}          & \textbf{1.85}     & \textbf{0.155} \\ \hline
\end{tabular}
}}
\vspace{-0.1cm}
\caption{\small Comparing DP-dyn{[}S,$\sigma${]} with adaptive clipping in terms of accuracy performance and time cost, with $C=4$ and $\sigma=6$.}
\label{table:adaptive_clipping}
\vspace{-0.4cm}
\end{table}

\section{Related Work}
 \label{sec:relatedwork}

 We have given an overview of related work in privacy threats and deep learning with differential privacy in Section~\ref{sec:introduction}. In this section, we focus on the most relevant literature. The first proposal for deep learning with DP~\cite{abadi2016deep} has been deployed in google TensorFlow~\cite{tfdldpimplementation}. However, one known problem for deep learning with DP is the degradation of model accuracy compared to the non-DP trained model. Several recent efforts have been put forward for improving the accuracy of the approach proposed in~\cite{abadi2016deep}. The most relevant efforts to this paper include the recent zCDP proposal with dynamic privacy budget allocation~\cite{yu2019differentially} instead of uniformed privacy budget allocation in~\cite{abadi2016deep}, and the adaptive clipping proposals represented by AdaClip~\cite{pichapati2019adaclip} and Quantile Clipping~\cite{thakkar2019differentially}. The main contribution of Yu et.al~\cite{yu2019differentially} is the new privacy accounting method zCDP that can compute the privacy spending when the DP training is using an approximate differential privacy under CDP with parameter $\rho$ to control dynamic privacy budget instead of $\epsilon$ privacy budget as in~\cite{abadi2016deep}. In addition, \cite{yu2019differentially} also proposed a dynamic privacy budget allocation solution to improve~\cite{abadi2016deep} which uses fixed privacy budget in every iteration of the DNN training, aiming to improving the accuracy of trained DNN model with DP. \cite{yu2019differentially} did not consider gradient leakage attack and resilience, and also uses the fixed clipping as the approximation to sensitivity in the same way as~\cite{abadi2016deep}. AdaClip performs the clipping bound estimation based on the coordinates of the gradient norm  and adaptively add different noise levels to different dimensions of the gradients. Quantile clipping is an alternative approach to AdaClip. It estimates the clipping bound during the training iterations using the quantile of the unclipped gradient norm instead of estimation of clipping bound based on the coordinates of the gradient norm in AdaClip. Both approaches are costly to compute dynamic clipping bounds since both needs to compute the clipping estimation on all M layers of a DNN for every example. Also both use the dynamic clipping bound to approximate the sensitivity while maintaining the fixed noise scale throughout the iterations of training to ensure that a sufficient amount of DP-noise is added according to the DP theory. Both are designed to improve accuracy of DNN training with DP but fail to be resilient against gradient leakages.
In comparison, our approach is by design both gradient leakage resilient and improving model accuracy thanks to injecting dynamic controlled DP noise. Our dynamic DP parameter optimization approach DP-dyn[$l_2$-max] incorporates dynamic sensitivity and dynamic noise scale to support decaying noise variance, such that as learning progresses in iterations, we inject less amount of DP controlled noises to the gradients instead of constant noise amount as done in~\cite{abadi2016deep}.  As a result, our dynamic $l_2$-max sensitivity provide a tighter DP controlled noise bound than the fixed Gaussian noise variance in used in conventional approaches.

\section{Conclusions}
\label{sec:conclusion}

We have presented a suite of algorithms with dynamic privacy parameters for gradient leakage resilient deep learning with differential privacy.
We first analyze some limitations of existing algorithms using fixed-parameter strategies that inject constant differential privacy noise to all layers during each training iteration. We then presented a suite of DP algorithms with dynamic parameter optimizations, including dynamic sensitivity mechanisms, dynamic noise scale mechanisms, and different combinations of dynamic parameter strategies. Extensive experiments on six benchmark datasets demonstrate that the proposed differentially private deep learning with dynamic hybrid sensitivity and dynamic decaying noise scale can outperform existing state-of-the-art approaches and other dynamic parameter alternatives with competitive accuracy performance, strong differential privacy guarantee, high resilience against gradient privacy leakage.

\vspace{0.6cm}

\textbf{Acknowledgement.}
The authors would like to first thank the associate editor Dr. Grigorios Loukides and the reviewers for their constructive and helpful comments. The authors acknowledge partial support by the National Science Foundation under Grants NSF 2038029, NSF 1564097, and an IBM faculty award.

\bibliographystyle{IEEEtran}
\bibliography{bare_jrnl.bib}





\end{document}